\documentclass[10pt]{article} %
\ifdefined\TMLRSubmission
  \usepackage{tmlr}
\else
  \usepackage[preprint]{tmlr}
\fi

\usepackage{microtype}
\usepackage{amsmath,amssymb,amsfonts}
\usepackage{bm}
\usepackage{graphicx}
\usepackage{placeins}
\usepackage{booktabs}
\usepackage{multirow}
\usepackage{caption}
\usepackage{subcaption}
\usepackage{xcolor}
\usepackage[colorlinks=true,linkcolor=blue!60!black,citecolor=blue!60!black,urlcolor=blue!60!black]{hyperref}
\usepackage{url}
\usepackage{tikz}

\title{LSTM-Style Looped Transformers with Shared KV Caches}

\ifdefined\TMLRSubmission
  \author{Anonymous authors}
  \hypersetup{pdfauthor={},pdftitle={LSTM-Style Looped Transformers with Shared KV Caches}}
\else
  \author{\name Andreas Kirsch \email blackhc@gmail.com}
  \hypersetup{pdfauthor={Andreas Kirsch},pdftitle={LSTM-Style Looped Transformers with Shared KV Caches}}
\fi

\def\month{10}
\def\year{2026}
\def\openreview{}


\begin{document}
\raggedbottom
\maketitle

\begin{abstract}
Recurrent transformers reuse weights to perform deep computation with fewer
parameters. We study an LSTM-style transformer that also shares one key--value
cache across its recurrent core. At width 1024, it uses 40.4\% fewer parameters
than an untied control at nearly matched forward FLOPs, and reduces persistent
cache sets from fifteen to four. Parallel training supplies cache entries from
a standard-attention pass, whereas continuation reads entries produced
recursively using shared attention. We address this mismatch by aligning the
two training passes' cache entries with a consistency loss during the final
10\% of a fixed training-token budget. Across three model widths on FineWeb,
this removes 81--89\% of the continuation-loss penalty relative to full-cache
evaluation. The improvement replicates on disjoint documents for two training
seeds. Across four widths and 1.05--4.19B training tokens, the loss cost of weight
tying is smallest at lower token budgets, suggesting an opportunity to use
recurrent computation when data and model storage are limited.
\end{abstract}

\section{Introduction}
\label{sec:intro}

Applying the same transformer blocks repeatedly separates parameter count from
computational depth. Universal Transformers, ALBERT, and recurrent language models
use this principle to share parameters across layers or iterations
\citep{dehghani2019universal,lan2020albert,geiping2025huginn}.
The corresponding attention caches are usually not shared: at each iteration, a token
attends to representations of earlier tokens from that same iteration. A direct
decoder therefore stores keys and values for every block application, even when its
weights are reused.

Shared attention memories already reduce this storage in both untied and looped
models \citep{wu2024lckv,sun2024yoco,conchello2026melt,oneill2026lla}.
We study the loss and computation costs of combining weight tying with one cache
for the entire recurrent core. The central question is whether a parallel training
objective prepares the model to reuse entries produced recursively during inference.

Our default model places four distinct transformer blocks between one untied
prologue block and two epilogue blocks. An LSTM-style update
\citep{hochreiter1997lstm,gers2000forget} combines each core application with a
carried state through input and forget gates, then normalizes the state before its
next use. Three iterations give fifteen block applications but only seven sets of
block weights. All core blocks read the final block's final-iteration cache for
preceding tokens (Figure~\ref{fig:flagship}).

\begin{figure}[t]
  \centering
  \begin{subfigure}[t]{\linewidth}
    \centering
    \resizebox{\linewidth}{!}{\begin{tikzpicture}[
    x=1cm, y=1cm, >=stealth, line width=0.55pt,
    font=\fontsize{9}{10.5}\selectfont,
    block/.style={draw=black!55, fill=black!4, rounded corners=1.8pt,
                  minimum height=0.74cm, inner sep=2pt, align=center},
    core/.style={block, fill=black!12, minimum width=0.55cm},
    flow/.style={->, draw=black!70},
    note/.style={align=center, inner sep=1.5pt},
    cache/.style={draw=accent!80, fill=accent!6, rounded corners=2pt,
                  minimum height=0.68cm, align=center, inner sep=3pt}]
\definecolor{accent}{HTML}{007E87}
\node[block, minimum width=1.1cm] (embed) at (0.45,0) {token\\embed.};
\node[block, minimum width=1.1cm] (pro) at (1.90,0) {prologue\\1 block};
\coordinate (hin) at (2.73,0);
\node[core] (f1) at (3.10,0) {$F_1$};
\node[core] (f2) at (3.81,0) {$F_2$};
\node[core] (f3) at (4.52,0) {$F_3$};
\node[core] (f4) at (5.23,0) {$F_4$};
\node[block, minimum width=0.75cm] (gnorm) at (6.27,0) {$N_g$};
\node[block, minimum width=3.15cm, minimum height=0.90cm] (cell) at (8.52,0)
     {LSTM-style cell\\[2pt]
      $c_r=f_r\odot c_{r-1}+i_r\odot g_r$};
\node[block, minimum width=0.75cm] (hnorm) at (10.95,0) {$N$};
\node[block, minimum width=1.04cm] (epi) at (12.10,0) {epilogue\\2 blocks};
\node[block, minimum width=1.30cm] (head) at (13.73,0) {output\\head};
\draw[flow] (embed) -- (pro);
\draw[flow] (pro) -- (f1);
\draw[flow] (f1) -- (f2);
\draw[flow] (f2) -- (f3);
\draw[flow] (f3) -- (f4);
\draw[flow] (f4) -- (gnorm);
\draw[flow] (gnorm) -- (cell);
\node[note, anchor=south] at (6.69,0.47) {$g_r$};
\draw[flow] (cell) -- (hnorm);
\draw[flow] (hnorm) -- (epi);
\node[note, anchor=west] at (11.40,0.66) {$h_r$};
\draw[flow] (epi) -- (head);
\draw[draw=accent, rounded corners=2pt]
     (11.39,0) -- (11.39,1.30) -- (2.73,1.30);
\draw[flow, draw=accent] (2.73,1.30) -- (hin);
\node[note, text=accent, fill=white] at (7.00,1.30)
     {three passes: reuse each $F_j$ and the gates};
\draw[draw=black!65] (10.25,0) -- (10.25,0.76) -- (8.47,0.76);
\draw[flow, draw=black!65] (8.47,0.76) -- (cell.north);
\node[note, fill=white, anchor=south] at (9.09,0.76)
     {carry to next pass};
\draw[draw=black!55, dashed] (hin) -- (2.73,-0.87) -- (8.47,-0.87);
\draw[flow, draw=black!55, dashed] (8.47,-0.87) -- (cell);
\node[note, fill=white] at (5.65,-0.87)
     {gates $i_r,f_r$ depend on $h_{r-1}$};
\node[note, text=black!65] at (11.38,-0.87)
     {$h_0=c_0=s$};
\node[note, text=black!65] at (2.32,-1.25)
     {$s=N(\text{prologue output})$};
\node[cache, minimum width=14.3cm] at (7.13,-1.91)
     {\textbf{One core KV cache, shared across all slots and passes}\\
      $F_4$'s final-pass K/V $\longrightarrow$ persistent cache
      $\longrightarrow$ earlier-token attention in $F_1,F_2,F_3,F_4$};
\node[note, text=black!65] at (7.13,-2.55)
     {Each block uses its own current-token K/V. Prologue and epilogue caches stay separate.};
\node[note, text=black!65] at (10.0,-1.25)
     {$N_g$: learned RMSNorm; $N$: gain-free RMSNorm};
\end{tikzpicture}}
    \caption{LSTM-style recurrence with one shared core cache.}
    \label{fig:arch}
  \end{subfigure}\\[5pt]
  \begin{subfigure}[t]{\linewidth}
    \centering
    \includegraphics[width=\linewidth]{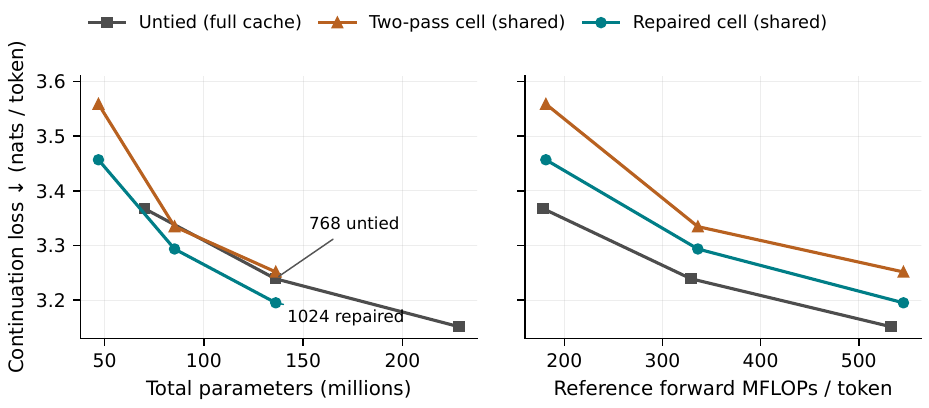}
    \caption{Continuation loss versus parameter count and forward computation.}
    \label{fig:scaling}
  \end{subfigure}
  \caption{\textbf{Parameter and cache sharing have distinct costs.}
  (a) The four core blocks have distinct weights, reused across iterations; preceding
  tokens supply one common donor cache. Prologue and epilogue retain their own caches.
  (b) Widths 512, 768, and 1024 after 4.19B training tokens, evaluated on the same
  1,280 document continuations with 2,048 prompt tokens and 512 scored targets.
  Repaired models add donor consistency during the final training stage.
  The FLOPs panel estimates one inference forward, not latency or the two-pass training
  cost (Appendix~\ref{app:flops}). All points use one training seed.}
  \label{fig:flagship}
\end{figure}

\begin{table}[t]
  \centering
  \caption{\textbf{Continuation loss, model size, forward computation, and persistent
  KV storage after 4.19B training tokens.} Full-cache and shared-cache losses score
  identical targets under ordinary prompt prefill (Section~\ref{sec:setup}).
  Repaired models repeat their parents' final training stage with donor consistency,
  retaining the same token budget. FLOPs use the reference attention lengths in
  Appendix~\ref{app:flops}; two-pass training costs about twice as much at fixed width
  and tokens.
  Sets and KV storage describe each model's serving layout: shared-cache columns
  use four sets; recurrent full-cache evaluation requires fifteen.
  BF16 KV storage is calculated from tensor dimensions for equal-length cache sets,
  excluding scratch space; window-sized allocations are discussed in
  Section~\ref{sec:memory}. Lower loss is better.}
  \label{tab:headline}
  \small
  \setlength{\tabcolsep}{4pt}
  \begin{tabular}{rlrrrrrr}
    \toprule
    Width & Model & \shortstack{Params\\(M)} & \shortstack{Forward FLOPs\\(M/token)} & Sets & \shortstack{KV\\(KiB/token)} & \shortstack{Full-cache\\loss $\downarrow$} & \shortstack{Shared-cache\\loss $\downarrow$} \\
    \midrule
    512 & Untied & 70.01 & 178.0 & 15 & 15 & 3.3679 & --- \\
        & Two-pass & 46.93 & 181.1 & 4 & 4 & 3.4395 & 3.5589 \\
        & Repaired & 46.93 & 181.1 & 4 & 4 & 3.4420 & 3.4570 \\
    \midrule
    768 & Untied & 135.98 & 328.9 & 15 & 22.5 & 3.2393 & --- \\
        & Two-pass & 85.24 & 336.0 & 4 & 6 & 3.2827 & 3.3347 \\
        & Repaired & 85.24 & 336.0 & 4 & 6 & 3.2839 & 3.2938 \\
    \midrule
    1024 & Untied & 228.49 & 532.9 & 15 & 30 & 3.1515 & --- \\
         & Two-pass & 136.20 & 545.5 & 4 & 8 & 3.1867 & 3.2518 \\
         & Repaired & 136.20 & 545.5 & 4 & 8 & 3.1878 & 3.1953 \\
    \bottomrule
  \end{tabular}
\end{table}

We use a common framework derived from modded-nanogpt
\citep{jordan2024moddednanogpt}\footnote{\url{https://github.com/KellerJordan/modded-nanogpt}}
to compare untied,
standard-cache recurrent, and shared-cache recurrent models across four widths and
three token budgets. Each width has a 15-layer untied control that approximately
matches the recurrent model's forward computation. Weight tying reduces parameters
at this fixed computational depth, with higher loss under the tested recipe.
A separate comparison pairs recurrent width 1024 with untied width 768 to
approximately match parameter count; the wider recurrent model achieves lower
continuation loss using more forward and training computation
(Figure~\ref{fig:flagship}, Table~\ref{tab:headline}).

Cache sharing introduces a second tradeoff. A two-pass objective can train all core
blocks to read a common cache, but its parallel second pass reads caches produced
by standard attention. During continuation, newly appended entries are themselves
produced using shared attention. We measure the total penalty of switching cache
policies on matched targets and examine the role of cache histories separately.
Adapting the KV-consistency idea of \citet{wu2024lckv}, we encourage the two passes
to write similar donor keys and values during the final training stage.
The intervention improves shared-cache continuation at widths 512--1024,
with a small full-cache loss cost, and its effect replicates on disjoint documents
for two width-768 training seeds.

Finally, we compare recurrent update rules and inference depths. Gated carry improves
loss at the trained depth in a controlled ablation; a Huginn-style injection baseline
also supports additional depth when trained across depths. Variable-depth training
and cache consistency can be combined, but the tested state-based stopping rule does
not beat fixed depth four. We therefore evaluate parameter efficiency, cache storage,
and adaptive computation separately.

\section{Method}
\label{sec:method}

\subsection{Backbone and gated recurrence}
\label{sec:backbone}\label{sec:cell}

An untied prologue $\mathcal P$ produces a seed representation, a core of $K$
distinct transformer blocks $F=F_K\circ\cdots\circ F_1$ is applied $R$ times, and an
untied epilogue $\mathcal E$ produces the final representation. We write P$i$E$j$
for $i$ prologue and $j$ epilogue blocks. Our default is P1E2 with $K=4$ and $R=3$:
seven sets of block weights and fifteen block applications. A \emph{slot} denotes a
block position in the core; weights are shared across iterations, not across slots.

Each block uses pre-RMSNorm with learned gains \citep{zhang2019rmsnorm}, rotary
positions \citep{su2021roformer}, grouped-query attention \citep{ainslie2023gqa},
and a SwiGLU feedforward layer \citep{shazeer2020glu}. We use an RMS variant of
query/key normalization \citep{henry2020qknorm}: gain-free RMS normalization within
each head, with fixed attention-logit scale $d_{\rm head}^{-1/2}$.
Head dimension is 64, with two query heads per KV head. Long and short attention windows
contain 2,048 and 1,024 tokens. The prologue uses a long window, the core repeats
long/short/short/short, and the epilogue uses short/long. The 15-layer untied control
has five long-window layers, at zero-based indices 0, 4, 7, 10, and 14, and ten short
ones. Both models therefore have the same counts of long and short block
applications. Input embeddings and the output projection share weights.
Appendix~\ref{app:recipe} gives remaining dimensions and initialization details.

Let $\mathcal N$ denote gain-free RMS normalization, and $\mathcal N_g$ the core's
learned-gain candidate normalization. For token embeddings $x$, initialize
$s=\mathcal N(\mathcal P(x))$ and $c_0=h_0=s$. One iteration computes
\begin{align}
  g_r &= \mathcal N_g(F(h_{r-1})), \label{eq:cand}\\
  f_r &= \sigma(W_fh_{r-1}+b_f),\qquad
  i_r = \sigma(W_ih_{r-1}+b_i), \label{eq:gates}\\
  c_r &= f_r\odot c_{r-1}+i_r\odot g_r, \label{eq:cell}\\
  h_r &= \mathcal N(c_r). \label{eq:hidden}
\end{align}
The gates act per token and coordinate. This is an LSTM-style update over depth,
without a $\tanh$ or output gate. The carry $c_r$ accumulates information; its
normalized read $h_r$ supplies the next core application, gates, and eventually the
epilogue. A separate learned-gain normalization precedes the output projection.
The candidate's learned gain can change its magnitude, while normalization of the
carry read controls the input scale of the next iteration.

Gate matrices start at zero, with $b_f=1$ and $b_i=0$. The initial gates are therefore
approximately 0.7311 and 0.5. With identity-initialized blocks, unit candidate gain,
and idealized exact normalization, the first update gives $c_1=1.2311s$ and
$h_1=h_0$. This preserves direction at initialization; the forget bias specifies
initial carry retention, not contraction of the complete state-dependent recurrence.

\subsection{One cache for all core blocks}
\label{sec:twopass}

Standard attention uses preceding-token keys and values from the same block and
iteration. A direct decoder stores $P+KR+E$ cache sets. Our shared-cache policy uses
one donor for the entire core: the own keys and values computed by the final core
block on its final iteration. Every core block, on every iteration, reads this donor
for preceding positions and its own key and value at the current position. Attention
still respects each block's window and document boundaries. Prologue and epilogue
blocks retain separate caches, giving $P+1+E$ sets: four instead of fifteen for the
default model. Keys are stored after query/key normalization and rotary positions.
This adapts final-layer KV sharing \citep{wu2024lckv} to a tied core while retaining
each block's own current-token KV.

The two-pass objective trains these reads in parallel over token positions. First,
run standard attention, obtain cross-entropy $\mathrm{CE}_1$, and retain the donor.
Second, restart the recurrence from the same prologue seed, use the retained donor
for shared-cache attention, and recompute the epilogue and $\mathrm{CE}_2$. We minimize
\begin{equation}
  \mathcal L_{\mathrm{2p}}=\tfrac12\mathrm{CE}_1+\tfrac12\mathrm{CE}_2.
  \label{eq:dual}
\end{equation}
Both terms are mean next-token cross-entropies. The donor remains in the computation
graph, so the second loss also updates the first-pass cache producer.

This construction does not make the second pass identical to autoregressive
continuation. In training, the donor for every preceding position was computed by
standard attention. During continuation, a newly processed token reads shared past
caches and appends its own final donor entry; subsequent tokens then read that entry.
Epilogue caches also develop recursively. Ground-truth token inputs do not remove
this difference in hidden-state histories. Comparing full-cache and shared-cache
continuations on identical targets measures the total effect of changing cache
policy, including both the read layout and the histories supplying it.

\subsection{Donor consistency during the final training stage}
\label{sec:consistency}

We adapt the successive-iteration KV MSE explored by
\citet[Appendix C.3]{wu2024lckv}, using a scale-normalized KV loss during the final
training stage. Let $X_1$ and $X_2$ denote the first and second passes'
own final donor tensor, for $X\in\{K,V\}$. Define
\begin{align}
 e_X &= \frac{\operatorname{mean}\big[(X_2-\operatorname{stopgrad}(X_1))^2\big]}
                    {s_X+10^{-6}}, \label{eq:kv-error}\\
 \mathcal L &= \mathcal L_{\mathrm{2p}}+
                   \tfrac{\gamma}{2}(e_K+e_V),\qquad \gamma=0.3.
 \label{eq:consistency}
\end{align}
The mean is over token positions, KV heads, and head coordinates. Each $s_X$ is a
reference first-pass mean square calibrated once at branch start and held fixed.
The recorded fixed-depth branches use 16 chunks of 8,192 tokens for calibration.
The auxiliary target is detached; the first-pass donor's gradient path through the
second cross-entropy remains live.

We apply this objective by branching from the start of the parent's final decay and
repeating its last 1,600 updates, preserving the optimizer state, token stream, and
schedule. Thus the repaired model has the same 4.19B-token budget as its parent;
10\% of its updates use consistency. This targets the stored donor, without directly
matching output probabilities or epilogue entries. Comparisons with those additional
terms and with whole-run consistency appear in Appendix~\ref{app:consistency}.
The loss encourages local agreement under first-pass histories; it does not enforce
agreement under every recursively generated history.

\subsection{Variable depth and stopping}
\label{sec:exit}\label{sec:uniformdepth}

For variable-depth training, each update draws one rank-synchronized
$R\sim\mathcal U\{3,\ldots,6\}$. At inference a token may stop when consecutive read
states point in similar directions. With the same gain-free normalization as above,
the implementation computes
\begin{equation}
 q_{r,t}=\frac{1}{d}\left\langle\mathcal N(h_{r,t}),
                                      \mathcal N(h_{r-1,t})\right\rangle.
 \label{eq:cos}
\end{equation}
This is a cosine up to normalization epsilon and finite precision. The token stops
at the first $r\ge2$ with $q_{r,t}>0.98$, or at the six-pass cap. Its selected stopping
iteration supplies its donor entry. Read-state direction, carry magnitude, and
prediction loss need not converge together.

The batched decoder freezes stopped states but continues computing until all rows
have stopped. We report each token's selected depth, not executed FLOPs or latency
savings. Parallel chunk probes instead use a global state statistic and a common
stopping depth. The adaptive consistency branch repeats the last 1,000 of 8,000
updates (12.5\%): 200 constant-rate updates followed by 800 decay updates.

\section{Experimental setup}
\label{sec:setup}

\paragraph{Training and comparisons.}
We train on FineWeb \citep{penedo2024fineweb} using a common framework derived from
modded-nanogpt \citep{jordan2024moddednanogpt}. Tied and untied models use the same
block components, data loader, and optimizer implementation. We simplify the
architecture to support controlled recurrence experiments.
We distinguish four trained variants: an \emph{untied} 15-layer control; a
\emph{standard cell} trained with single-pass cross-entropy and separate caches;
a \emph{two-pass cell} trained with Equation~\ref{eq:dual}; and its
\emph{repaired} branch trained with Equation~\ref{eq:consistency} near the end.
The standard cell is a separate model, not the first pass of the two-pass cell.

The main ladder uses widths 384, 512, 768, and 1024 and fixed depth three.
Endpoints of 4,000, 8,000, and 16,000 updates contain 1.048576B, 2.097152B, and
4.194304B training tokens. Each update processes 262,144 tokens in 32 chunks of
8,192, with attention confined to chunks and documents. Training uses BF16 on eight
H100s, global gradient clipping at norm 1, Muon for hidden matrices
\citep{jordan2024muon}, and AdamW \citep{loshchilov2019adamw} for embeddings,
normalization gains, and biases.
The selected learning rates are 0.01 and $3\times10^{-4}$, respectively. After an
80-update warmup, a constant-rate trunk supports branches with linear decay over the
last 10\% of each budget (Appendix~\ref{app:recipe}). Common named tensors share
initial values in paired controls. Training data order is fixed across seeds.
The repository maps models to recorded configurations and results.

\paragraph{Chunk validation.}
All losses are next-token cross-entropy in nats. Standard and parallel shared-cache
validation score the same 1,280 chunks of 8,192 targets (10,485,760 targets total).
Standard validation uses each block and iteration's own past representations.
For the main two-pass models, parallel shared-cache validation is the second-pass computation of
Section~\ref{sec:twopass}. These provide the model-size ladder and update-rule
comparisons. They do not measure a cache that is updated recursively during
continuation.

\paragraph{Prompt-prefill continuation.}
Our primary continuation evaluation uses 1,280 document-contained spans, each with 2,048
prompt tokens followed by 512 scored targets (655,360 targets total). Ordinary
prefill computes the prompt with standard attention. The full-cache reference
retains all block-iteration caches; the shared-cache continuation retains the final
donor and prologue/epilogue caches. Subsequent token processing updates the respective
caches recursively. Both score the same targets and always receive ground-truth
token inputs. The first target uses the final prompt logits, followed by 511 decoding
steps. We call the shared-cache result \emph{served loss}; it measures teacher-forced
continuation, not sampled generation quality.

A 512-prompt/2,048-target diagnostic changes both continuation length and target
positions. A two-pass-prefill diagnostic retains first-pass donor entries but
second-pass epilogue entries and logits. \emph{Cold} evaluation processes both prompt
and continuation sequentially from empty caches under the shared policy; it is
reported separately. After selecting the repair, we evaluate it on the next 1,280 eligible
documents disjoint from the original spans and chunk set; this is fresh same-source
evaluation, not a new training replicate (Appendix~\ref{app:evaluation}).
The appendix's update-rule and cache-producer pilots use separate held-out
documents and state their evaluation policy alongside each result.

\paragraph{Resources and uncertainty.}
Parameter counts include tied embeddings once and all used weights. Reference forward
FLOPs count dense projections and attention products for one pass at the stated depth,
using mean attended lengths in an 8,192-token chunk with 2,048/1,024 windows.
They omit elementwise operations and are not measured serving latency. We also
report estimated training FLOPs because two-pass and single-pass objectives differ
substantially in cost (Appendix~\ref{app:flops}). KV storage counts persistent BF16
keys and values, excluding scratch space.

Width-768 untied and two-pass ladders have three training seeds; repair confirmation
uses two of them. Other widths and update-rule/deep-depth experiments use one seed.
Paired continuation errors use a block bootstrap over ordered document spans, with
blocks of eight and 2,000 resamples. These conditional standard errors describe
variation over evaluation documents, not over training runs. We report training-seed
variation separately (Appendix~\ref{app:seeds}); the compact prespecification summary
is in Appendix~\ref{app:prereg}.

\section{Results}
\label{sec:results}
\subsection{Model size and computation}
\label{sec:params}\label{sec:family}

\begin{figure}[t]
  \centering
  \includegraphics[width=\linewidth]{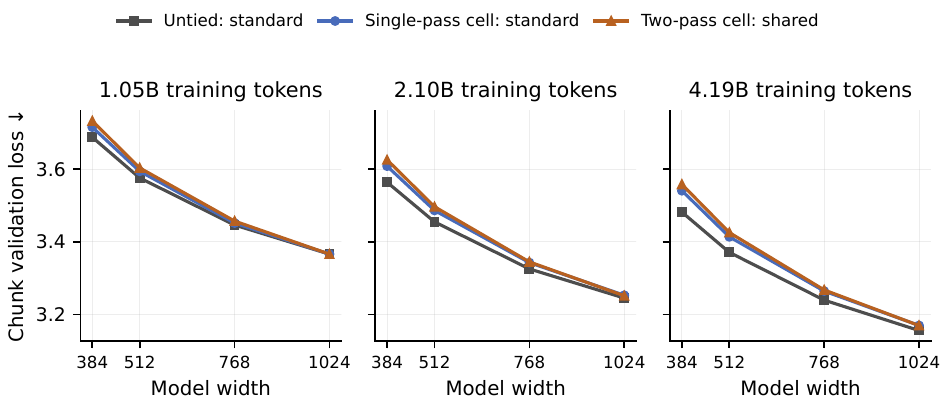}
  \caption{\textbf{Loss across widths and training-token budgets.}
  The three models are separately trained. Standard and parallel shared-cache
  validation score the same 1,280 chunks (10,485,760 targets). Each width has a
  15-layer untied control approximately matched to one recurrent forward;
  two-pass training costs more. Points use seed 0; Appendix~\ref{app:seeds}
  repeats the width-768 two-pass/untied comparison over three seeds.
  Appendix~\ref{app:ladder} replots the losses against reference forward FLOPs.
  These parallel losses do not measure shared-cache continuation.}
  \label{fig:ladder}
\end{figure}

The primary controls match width, training tokens, and the number of block
applications. Their reference forward FLOPs differ by only 1.5--2.4\%, from
the recurrent gates. Untied models have lower loss at each width
(Figure~\ref{fig:ladder}). We also compare models of different widths with
approximately equal parameter counts (Figure~\ref{fig:flagship}).

\subsubsection{Fixed width and training tokens}
\label{sec:flops}\label{sec:tokens}

At 4.19B tokens, both recurrent families have higher loss than the untied control
at every width, with smaller gaps for wider models (Figure~\ref{fig:ladder}).
At width 1024 and 1.05B tokens, single-pass standard and two-pass parallel shared
losses are within 0.0002 nats of the untied control with 40.4\% fewer parameters
(seed 0; Table~\ref{tab:app-ladder}).

All model families improve from 1.05B to 4.19B tokens, while both recurrent families'
gaps to the untied controls increase at every tested width, despite nearly matched
forward computation (Figure~\ref{fig:app-ladder-flops}). The loss cost of reducing
parameters through recurrence is therefore smaller at the lower token budgets.
Across three training
seeds at width 768, the two-pass model's shared-minus-untied gap grows from
0.0115 to 0.0209 to 0.0309, with seed standard deviations 0.0009, 0.0014
and 0.0025, respectively. This supports the budget trend at that width; the
other widths and the single-pass cells have one seed each. These observations
do not identify an asymptotic capacity limit or a width at which parity would
be reached.

Matching estimated training compute instead pairs the two-pass models at 2.10B
tokens with untied controls at 4.19B. Their parallel shared-cache losses are higher
by 0.0957--0.1439 nats across the four widths
(Table~\ref{tab:app-training-compute}). These comparisons precede consistency repair
and do not measure recursive continuation quality.

\subsubsection{Cross-width comparison at approximately matched parameter count}

After donor-consistency training, the width-1024 recurrent model has 136.202M
parameters, close to the width-768 untied model's 135.978M. On the same primary
continuation spans their losses are 3.1953 and 3.2393: a paired difference of
$-0.0441$ nats (block-bootstrap SE 0.0014). The recurrent model uses
545.5M rather than 328.9M reference forward FLOPs per token, a factor of 1.66.
Its estimated training computation is 3.22 times larger at the same token
budget, excluding consistency-loss overhead. This cross-width comparison matches
parameters, not forward computation. The width-1024 untied compute control has
228.49M parameters and uses 532.9M reference forward FLOPs per token; its loss
is 3.1515, compared with the recurrent model's 3.1953
(Table~\ref{tab:headline}). The cross-width comparison uses seed 0
on the original spans; it has not received the held-out confirmation used
for the width-768 repair in Section~\ref{sec:cache}. Detailed parameter and
compute accounting is in Appendix~\ref{app:flops}.

\subsection{Update-rule controls and a Huginn-style baseline}
\label{sec:cellvs}\label{sec:arms}

We compare the gated-carry update with a plain update that removes the carry and
gates and overwrites the hidden state with the normalized candidate,
$h_r=\mathcal N(g_r)$. For each update, we also test input injection: a learned
linear adapter projects the concatenation of the current hidden state and the
prologue state $s$, giving core input $A[h_{r-1};s]$ at every iteration.
The four factorial arms in Table~\ref{tab:arms} share the backbone,
initialization of common parameters, data
order, three-iteration depth, single-pass objective and training schedule.
Removing the cell also removes carry accumulation, so this control does not
isolate learned gates from the carried residual state.

We additionally test a \emph{Huginn-style} baseline, adapting the input-injection
recurrence of \citet{geiping2025huginn} to our common backbone. It uses the same
adapter at every iteration, starts from a normalized Gaussian hidden state, and
sets $h_r=g_r$: the core's learned-gain candidate becomes the next hidden state
directly, without a cell carry or gates. This differs from the plain adapter arm,
which starts from $s$ and applies gain-free normalization to the candidate.
The fixed-depth comparison uses three iterations; Section~\ref{sec:probes}
separately tests variable-depth training and gradient truncation.
These controls use standard caches. A separate width-512 shared-cache pilot
tests the repair across update rules (Appendix~\ref{app:shared-update-controls}).

\begin{table}[htbp]
  \centering
  \caption{\textbf{Recurrent update and input-adapter controls.}
  P1E2 at width 768, three iterations and 1.05B training tokens, one training
  seed. Losses use standard attention on 10,485,760 targets.
  Parentheses are paired block-bootstrap SEs of the difference from the cell,
  conditional on these checkpoints. Parameter counts exclude embeddings.}
  \label{tab:arms}
  \small
  \setlength{\tabcolsep}{6pt}
  \begin{tabular}{llrrr}
    \toprule
    Update & Adapter & Params (M) & Loss $\downarrow$ & $\Delta$ loss $\downarrow$ \\
    \midrule
    Gated carry & No & 46.61 & 3.4517 & --- \\
    Plain overwrite & No & 45.43 & 3.4838 & $+0.0321$ (0.0003) \\
    Gated carry & Yes & 47.79 & 3.4518 & $+0.0001$ (0.0003) \\
    Plain overwrite & Yes & 46.61 & 3.4683 & $+0.0166$ (0.0003) \\
    Huginn-style & Yes & 46.61 & 3.4745 & $+0.0228$ (0.0003) \\
    \bottomrule
  \end{tabular}
\end{table}

The gated-carry update improves loss relative to plain overwrite at the trained
depth. Input injection reduces the plain model's loss by 0.0155 but changes the
cell's by only $+0.0001$, closing about half the plain-versus-cell gap.
Neither cell variant tolerates doubling the evaluation depth:
at six iterations their losses rise to 4.0125 and 4.1154, whereas the plain
adapter and Huginn-style arms reach 3.5690 and 3.5266. Low loss at the trained
depth and robustness to additional iterations are separate properties.

\subsubsection{Gradient truncation and additional depth}
\label{sec:huginn}\label{sec:probes}

At fixed depth three, the Huginn-style baseline remains 0.022--0.025 nats
above the single-pass cell as training increases from 1.05B to 4.19B tokens
(Table~\ref{tab:app-update-budget}). The other three update-rule ablations were
trained only to 1.05B tokens.
For a separate comparison at greater depth, we train both updates at width
512 over $R\sim\mathcal{U}\{4,\ldots,32\}$ on 1.05B tokens, with standard
caches and either full gradients or gradients through only the last eight
iterations. Token budgets and sampled depths are matched; training computation
differs between gradient treatments. At evaluation depth 32, full-gradient losses are 3.6065 for the
cell and 3.6036 for injection, a difference of $-0.0029$ (SE 0.0003).
This small gap concerns these checkpoints, not statistical equivalence across runs.

Truncation changes the comparison substantially: it raises the cell's loss
by 0.0757 and lowers injection's by 0.0140, producing a 0.0926 gap.
When the depth exceeds eight, truncation cuts the cell's only gradient path
to its prologue, while injection retains a path through the adapter on later
iterations. The controls establish an architecture-dependent effect of
truncation, not a general ranking of update rules. At depth 64, injection's
loss is almost unchanged while the full-gradient cell's rises by 0.0222.
The best deep arm reaches 3.5896 at depth 32, compared with 3.5941 for the
fixed-depth-three cell at the same width and token budget, using 6.5 times
its forward FLOPs. Full curves are in Appendix~\ref{app:depth}.

All these update-rule comparisons have one training seed. They adapt Huginn's
recurrent structure to a common smaller-scale language-model setting
\citep{geiping2025huginn}; they do not reproduce its scale, training recipe or
reasoning evaluations. The results therefore do not resolve whether its
reported benefits from additional inference depth transfer to this regime.

\subsection{Shared-cache continuation and donor consistency}
\label{sec:cache}\label{sec:slotshare}

A low parallel shared-cache loss does not ensure accurate continuation through
a recursively updated cache. At width 768, the two-pass model's shared loss
on chunks is 3.2675, below its standard loss of 3.2728. On the primary
continuation spans, however, its full-cache loss is 3.2827 and its shared-cache
continuation loss is 3.3347: a penalty of 0.0520 (SE 0.0013).
These are different evaluation sets, so the penalty is computed within the
continuation protocol, on identical targets. Token inputs remain ground truth;
the recursively produced quantities are cache entries, not sampled tokens.

\begin{figure}[t]
  \centering
  \includegraphics[width=\linewidth]{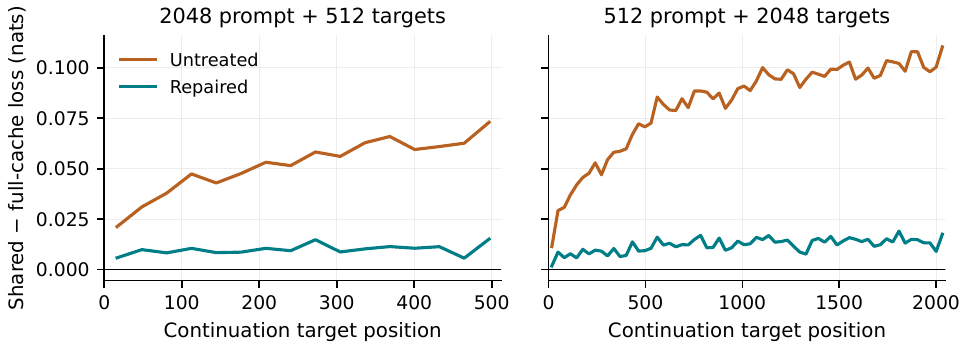}
  \caption{\textbf{Excess loss during shared-cache continuation.}
  Width 768, 4.19B training tokens, seed 0, before and after late donor
  consistency. Each curve subtracts the same checkpoint's full-cache loss
  at the same positions. Both settings use the original 1,280 documents;
  prompt and continuation lengths differ together. The repair reduces the
  excess loss throughout the continuation, without making the two cache
  computations equivalent.}
  \label{fig:cache-drift}
\end{figure}

The two-pass objective exposes each token to donor entries from a standard
first pass. During continuation, it instead reads entries written by earlier
shared-cache forwards. The continuation contrast changes both read layout and
cache history. Figure~\ref{fig:cache-drift} shows how this total penalty varies
with continuation position. The shorter-prompt,
longer-continuation diagnostic has a larger average penalty: 0.0826 at
width 768. That comparison changes both conditioning and scored positions,
so it does not isolate continuation length alone.

A separate diagnostic holds the shared-read policy fixed and repeatedly recomputes
donors on the same 64 chunks. From the second to third iterate, the parent's loss
rises from 3.2510 to 3.2994; after repair it rises from 3.2515 to 3.2596.
This supports a cache-history contribution, but does not decompose the
prompt-prefill continuation penalty (Appendix~\ref{app:donor-iterate}).

\subsubsection{Late consistency training}

Applying donor consistency during the final 10\% of training reduces the
continuation penalty at all three evaluated widths (Table~\ref{tab:repair}).
Across both continuation settings, 81--93\% of the within-model penalty is removed.
The continuation's full-cache loss increases by 0.0011--0.0025.
The separate prespecified guard on chunk-standard degradation also passes
its 0.005 tolerance.
The repaired models still trail their width-matched untied controls by
0.0891, 0.0545 and 0.0438 nats at widths 512, 768 and 1024
(Table~\ref{tab:headline}).

\begin{table}[t]
  \centering
  \caption{\textbf{Effect of late donor consistency.}
  Repaired minus untreated checkpoint at 4.19B tokens, seed 0, on the primary
  2,048-prompt/512-target setting. Parentheses are paired block-bootstrap SEs.
  The residual penalty subtracts the repaired model's own full-cache loss
  from its shared-cache continuation loss. Absolute losses and resources
  are in Table~\ref{tab:headline}.}
  \label{tab:repair}
  \small
  \begin{tabular}{rrrr}
    \toprule
    Width & $\Delta$ continuation loss $\downarrow$ & $\Delta$ full-cache loss $\downarrow$ & Residual penalty \\
    \midrule
    512 & $-0.1019$ (0.0022) & $+0.0025$ (0.0002) & $+0.0150$ \\
    768 & $-0.0409$ (0.0010) & $+0.0012$ (0.0003) & $+0.0099$ \\
    1024 & $-0.0565$ (0.0013) & $+0.0011$ (0.0003) & $+0.0075$ \\
    \bottomrule
  \end{tabular}
\end{table}

At width 768, a second training seed reproduces the improvement. On a further
1,280 disjoint documents reserved for confirmation, continuation losses
improve by 0.0434 (SE 0.0011) and 0.0298 (SE 0.0006) for seeds 0 and 1;
their full-cache costs are 0.0010 and 0.0014. The corresponding penalty
reductions are 80\% and 75\%. This confirms the selected repair on fresh
same-source documents, with different effect sizes across the two seeds;
it does not establish cross-domain generalization.

\paragraph{Loss variants.}
At width 768, increasing donor weight from $\gamma=0.1$ to 0.3 to 1.0 lowers
continuation loss from 3.3001 to 3.2938 to 3.2902, while full-cache costs
relative to the rerun control rise from 0.0005 to 0.0012 to 0.0057.
We select 0.3 using the chunk-standard quality guard: its cost is 0.0013,
whereas weight 1.0 costs 0.0052, just outside the 0.005 tolerance.
Adding output distillation to combined donor/epilogue consistency almost
eliminates the penalty but costs 0.0112 in chunk-standard loss. Donor-only
supervision achieves the improvement; its slight advantage over combined
supervision does not isolate epilogue effects, because the combined loss
halves the donor coefficient. Applying consistency throughout training
also changes its normalization from fixed to tracked scales; those comparisons
do not isolate the effect of timing. Appendix~\ref{app:consistency} reports
the variants and both continuation settings. A separate width-512 pilot also finds
that donor consistency improves plain and input-injection updates, so the repair
does not require the gated carry at this setting
(Appendix~\ref{app:shared-update-controls}).

\subsubsection{Cache storage and measured memory}
\label{sec:memory}

With two query heads per KV head, one BF16 cache set uses $2d$ bytes per token.
The shared layout has four sets rather than fifteen, giving 3.75 times less
storage with uniform-length rings, or 2.86 times less when each cache is
allocated to its actual attention window. In an eager decoder on one H100,
filled-cache peak memory at batch 160 is 2.77 and 2.72 times lower than the
same cell's standard layout at widths 768 and 1024; the ratios against the
untied models are 2.87 and 2.85. These measurements include implementation
workspace during sequential cache fill, excluding ordinary parallel-prefill
workspace, and are distinct from KV tensor sizes. They do not establish a
latency benefit: at width 1024 and batch 160, a shared-cache decode step takes
37.2\,ms versus 20.7\,ms with the cell's standard layout. The cause of this
difference has not been isolated. Appendix~\ref{app:hardware} gives the
batch-size measurements and their implementation limits.

\subsection{Variable depth and per-token stopping}
\label{sec:adaptive}\label{sec:dynamics}

Training over $R\sim\mathcal{U}\{3,\ldots,6\}$ makes loss less sensitive to
evaluation depth. At width 768 and 2.10B tokens, the variable-depth two-pass
model's standard chunk loss ranges from 3.3477 to 3.3551 over depths four
through eight. A model trained at depth three instead deteriorates from
3.3504 at depth three to 3.7543 at depth six. The robustness has a cost:
at depth three, the variable-depth model's standard loss is 0.0188 higher
than the fixed-depth model's. At depth four it recovers approximately the
fixed-depth model's depth-three loss, using an additional core pass.

\subsubsection{Shared-cache continuation}
\label{sec:decode}\label{sec:halting}

Donor consistency also improves the variable-depth model. We re-run its
last 1,000 of 8,000 updates with consistency enabled and fixed reference
scales. On the primary continuation spans, loss improves by 0.0325--0.0370
at fixed depths three, four and six and under per-token stopping, while
full-cache loss increases by 0.0011--0.0017. The residual penalties at
fixed depths three, four and six are $-0.0015$, $+0.0082$ and $+0.0136$.
This supports compatibility of the repair with variable-depth training,
but its effect is not uniform across evaluation depths.

\begin{figure}[t]
  \centering
  \includegraphics[width=0.84\linewidth]{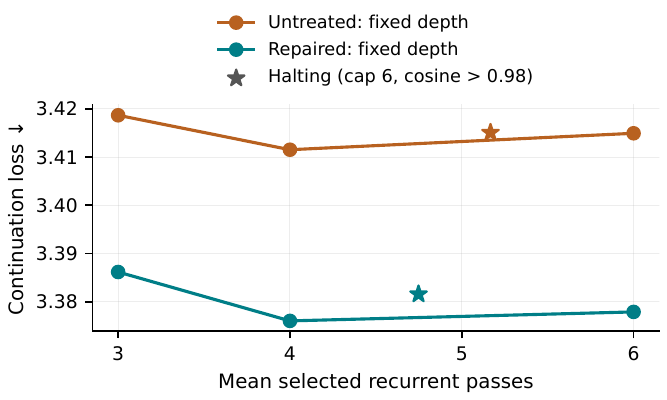}
  \caption{\textbf{Fixed depth outperforms the tested stopping rule.}
  The width-768 model is trained over depths 3--6 on 2.10B tokens, with and
  without late donor consistency. Losses use shared-cache continuation on
  the primary 1,280 spans. Fixed depths are three, four and six; per-token
  stopping uses successive-state cosine $\tau=0.98$, minimum depth two and
  cap six. The horizontal axis counts selected passes, not executed GPU
  computation. There is one training seed.}
  \label{fig:adaptive-depth}
\end{figure}

The tested stopping rule does not improve the loss--depth tradeoff
(Figure~\ref{fig:adaptive-depth}). For the repaired model, fixed depth four
gives loss 3.3761 at four passes. Stopping selects 4.75 passes on average
and gives loss 3.3816, a paired disadvantage of 0.0055 (SE 0.0002).
Fixed depth four also dominates stopping for the untreated model and for
the model trained with consistency from the start. With worse loss and more
selected passes, this stopping rule offers no advantage over that fixed-depth
control. Full depth and consistency comparisons are in Appendix~\ref{app:adaptive}.

\section{Related work}
\label{sec:related}

\paragraph{Cache sharing and consistency.}
Layer-Condensed KV Cache (LCKV) shares final-layer keys and values across layers
and handles their sequential dependency through iterative parallel cache refinement
\citep{wu2024lckv}. Its Appendix~C.3 also tests an MSE loss between successive cache
iterates, with lower perplexity in one configuration and higher perplexity in
others. We adapt this idea to an LSTM-style tied core using normalized donor
matching late in training. Our two-pass objective starts with standard attention,
and each block retains its own current-token K/V; LCKV masks that position.
Our two- and four-sweep controls vary refinement within this recipe, with full
gradients (Appendix~\ref{app:refinement-training}).

Feedback Transformer uses one shared memory of past-token layer representations and
shared K/V projections \citep{fan2020feedback}. Cross-Layer Attention reuses earlier
layers' KVs \citep{brandon2024cla}, and YOCO's cross-decoder shares caches produced by
its self-decoder \citep{sun2024yoco}. \citet{wu2025systematic} compare lower- and
upper-layer donors, including their training and prefill costs. Our early-donor
ablation tests this choice in a tied core (Appendix~\ref{app:early-producer}).
Huginn tests zero-shot sharing across iterations
\citep[Section~6.2]{geiping2025huginn}, and a Mixture-of-Recursions variant reuses
first-recursion KVs \citep{bae2025mor}. Our final donor is shared across both
iterations and distinct core blocks.\footnote{Concurrent work by
\citet{huang2026fixedpoints} studies terminal KV reuse across iterations,
learned training-depth priors, and orthogonal input injection. It retains one
cache per physical layer; we share one donor across distinct core blocks and
evaluate finite-depth donor consistency.}

MELT adapts pretrained Ouro models using gated cache-writing states, chunk-wise
training, and distillation at every loop \citep{conchello2026melt}. It shares a global
cache across loops for each layer while retaining small per-loop sliding-window
caches. Our gates update the recurrent carry, and all core blocks share one donor.
Looped Latent Attention compresses loop trajectories and reconstructs loop-specific
KVs using a distilled codec \citep{oneill2026lla}. Its unsuccessful final-loop-reuse
control leaves the original model frozen; our shared-cache models are trained for
that policy. FlashLoop uses sparse updates and quantized cross-loop KV residuals
without retraining \citep{yang2026flashloop}. These methods adapt pretrained models;
we study training from scratch and subsequent consistency repair.

\paragraph{Depth recurrence and gating.}
Universal Transformers apply shared updates with adaptive computation
\citep{dehghani2019universal}, ALBERT shares layer parameters \citep{lan2020albert},
and deep equilibrium models formulate prediction through a fixed point of a tied
transformation \citep{bai2019deq}. Our finite recurrence uses an LSTM-style carry
\citep{hochreiter1997lstm,gers2000forget}, with independent input and forget gates
and a normalized read. The controls test the combined carry/gate update.

Huginn combines a prelude/core/coda architecture with input injection, randomized
depth, and truncated backpropagation \citep{geiping2025huginn}. Our Huginn-style arm
adapts these mechanisms to a common backbone. Its full- and truncated-gradient
controls use standard caches; shared-cache update controls appear separately in
Appendix~\ref{app:shared-update-controls}. Neither reproduces Huginn's scale or
full recipe.
Relaxed Recursive Transformers convert pretrained models using layer-specific
low-rank adapters \citep{bae2025relaxed}. Mixture-of-Recursions learns token-level
depth routing \citep{bae2025mor}, and Ouro uses an entropy-regularized exit policy
\citep{zhu2025ouro}. Our stopping rule uses state agreement without learned routing.

\paragraph{Scaling recurrence.}
Parcae studies allocation between training tokens and loop depth at fixed parameter
count \citep{prairie2026parcae}. At matched executed depth,
\citet{schwethelm2026recurrence} find that their baseline recurrence saves parameters
but increases loss at matched training computation. SMELT estimates training-compute
savings at matched loss for a tuned looped mixture-of-experts recipe with closely
matched per-token FLOPs, parameters, and cache storage \citep{wang2026smelt}.
\citet{chen2026growth} distinguish weight sharing, boundary operators, and model
growth, finding gains from more recurrence when training repeatedly on limited data.

\paragraph{Distillation and adaptive computation.}
MiniLM matches attention distributions and value relations \citep{wang2020minilm};
our auxiliary target is the stored donor itself. Adaptive computation time and
PonderNet learn stopping policies \citep{graves2016act,banino2021pondernet}.
CALM addresses early exits and missing deeper past-token representations
\citep{schuster2022calm}, while LayerSkip uses intermediate exits for self-speculative
decoding \citep{elhoushi2024layerskip}. Our state-agreement rule has neither a learned
halting loss nor speculative verification and is evaluated against fixed-depth
controls under the same cache policy.

\section{Discussion and limitations}
\label{sec:discussion}

The loss penalty for tying grows with the training-token budget at nearly matched
forward computation, suggesting more scope for parameter savings when data are
limited. Our depth-three ladder extends the training stream as the budget grows.
On a 100M-token corpus reused over ten epochs, \citet{chen2026growth} find that
increasing recurrence can outperform adding parameters. Whether shared-cache models
retain those multi-epoch gains is open. Our grid neither optimizes width/depth nor
fits scaling exponents. All families improve through 4.19B tokens; we observe no
capacity floor or compute-optimal training endpoint.

When training is amortized over many predictions, loss at a given inference and
cache budget may justify more training. The cache-producer comparisons illustrate
the distinction: early production wins the pilot at equal training compute, while
an existing late-producer checkpoint has lower loss on fresh documents after about
four times as much estimated training compute, at nearly equal inference cost
(Appendix~\ref{app:early-producer}). Neither result establishes a generally better
producer or attributes the gap to recurrently refined memory alone.

Donor consistency reduces the shared-cache continuation penalty. The repeated-donor
diagnostic supports a contribution from cache histories, without decomposing the
full penalty. The improvement requires no direct epilogue supervision, but does not
exclude epilogue effects or guarantee agreement on other histories. The confounds
in the loss-variant controls remain (Appendix~\ref{app:consistency}). Two seeds
confirm the selected repair on fresh same-source
documents, without establishing a generally optimal coefficient or zero quality cost.

The repair also helps plain and injection updates in a single-seed pilot; the
tested LSTM-style carry is not necessary for that improvement. Four refinement
sweeps provide no advantage over two sweeps with late consistency at matched
estimated training compute (Appendix~\ref{app:refinement-training}); their equal-token
ranking depends on prompt history. These pilots do not establish a best update
rule or producer across scales. Holding normalization fixed would separately test
the value of applying consistency late.

Evaluation covers FineWeb language-model loss with ground-truth inputs. Sampled
generation, longer contexts, downstream tasks, and larger models may behave
differently. Most configurations have one run; document-level uncertainty excludes
training variability. Our small-scale Huginn-style experiments leave open how
architecture, data, and optimization affect the value of additional depth.

Cache storage and selected depth do not determine latency. Uniform cache rings
overstate savings relative to window-sized allocations; our implementation slows
at width 1024 with large batches. The adaptive decoder computes stopped rows until
the batch completes. A quality--latency improvement over fixed depth remains to
be demonstrated.

\section{Conclusion}
\label{sec:conclusion}

An LSTM-style weight-tied transformer reduces parameter count at fixed
computational depth and shares one persistent cache across its recurrent core.
Two-pass training leaves a shared-cache continuation penalty; late donor-KV
consistency substantially reduces it, with a small full-cache loss cost and
confirmation on disjoint documents.
At matched width and approximately matched forward computation, untied controls
retain lower loss. Simpler updates
also benefit from consistency in a pilot, and adaptive stopping does not improve
on fixed depth. The evidence supports cache sharing with a measured quality and
computation tradeoff, rather than a general advantage for gated recurrence.

\bibliographystyle{tmlr}
\bibliography{refs}

\clearpage
\appendix
\section{Acknowledgements}
\label{app:acknowledgements}

The experiments, code, and manuscript were developed using Fable 5.1 and
GPT-6 Astra. The author verified and edited the resulting work.

Experiments used a local workstation with an NVIDIA GeForce RTX 4090 and Modal
cloud instances. Main model training used eight NVIDIA H100 GPUs per job; brief
hardware-selection benchmarks also used H200 and B200 GPUs.

\section{Architecture and training recipe}
\label{app:recipe}
\label{app:conventions}
\label{app:training}

All models use the same transformer block: a learned-gain RMS normalization before
attention and before the MLP, rotary positional embeddings, gain-free per-head query/key
normalization, and a SwiGLU MLP. Attention has 64-dimensional heads and two query heads
per key/value head. For model width $d$, the MLP hidden width is $8d/3$ rounded up to a
multiple of 128. The tokenizer has 50,257 entries; input and output embeddings share
one matrix padded to 50,304 rows. There are no value-embedding tables, raw-embedding residual skips, learned
residual-mixing scalars, or FP8 output projection in these experiments.

The recurrent model has one prologue block, four distinct core blocks, and two epilogue
blocks (P1E2). The core weights are reused at each iteration. Its prologue, first core
slot, and last epilogue use a 2,048-token attention window; other blocks use 1,024 tokens.
At three iterations this gives five long-window and ten short-window block applications.
The untied comparator has 15 blocks, with long windows at zero-based indices
0, 4, 7, 10, and 14. Thus the models match the counts of long and short applications,
while placing them differently in the stack. Causal attention is also restricted to the
current document and logical chunk. The maximum attention window and the 8,192-token
logical chunk length are distinct quantities.

Attention and MLP output projections start at zero, so each transformer block initially
acts as the identity. Other projection matrices use a uniform distribution with standard
deviation $0.5/\sqrt{\mathrm{fan\_in}}$; embeddings have independent Gaussian entries with
standard deviation 0.02. Normalization gains start at one. Gate matrices start at zero,
with forget and input biases 1 and 0. The initial forget coefficient
$\sigma(1)\simeq0.731$ describes carry retention, not contraction of the full recurrent
map. The cell starts with hidden state and carry equal to the gain-free-normalized prologue output. Its carry
read is gain-free RMS normalization; the candidate and final output normalizations have
learned gains. There is no additive latent noise in the cell. The Huginn-style comparator
instead starts from a unit-RMS Gaussian hidden state, uses the prologue adapter on every
pass, and reads the normalized candidate directly.

\paragraph{Data and optimization.}
The global batch contains 262,144 tokens, arranged as 32 logical chunks of 8,192 tokens.
All paired models see the same ordered FineWeb training stream. An update's chunk indices
are determined by its step, independently of microbatching. Training uses eight H100 GPUs
with four chunks per rank per update. Memory-intensive runs split these into two or four
microbatches; accumulation preserves the global batch. Attention does not cross logical
chunk boundaries. Training does not cycle over exhausted data.

Hidden matrices use Muon with learning rate 0.01, momentum 0.95, five Newton--Schulz
iterations, weight decay 0.1, and update scaling
$\sqrt{\max(1,\mathrm{rows}/\mathrm{columns})}$. The tied embedding and remaining
parameters use AdamW at learning rate $3\times10^{-4}$, with
$\beta=(0.9,0.95)$ and $\epsilon=10^{-8}$. Embeddings have weight decay 0.1; normalization
gains and biases have none. Gradients are accumulated and synchronized before global
norm clipping at 1.0. The recipe is common across model widths, rather than separately
tuned to each architecture. Models compute in BF16 with FP32 normalization and loss
reductions. Optimization states and complete random-number-generator states are saved
for continuation. The ladder uses PyTorch 2.14 and CUDA 13.

Each main-ladder schedule warms up for 80 updates and then holds its learning rate constant.
Separate branches decay linearly to zero over the final 10\% of each endpoint
(Table~\ref{tab:app-schedule}). The constant-rate trunk continues independently of its
earlier decay branches. An early annealed endpoint is therefore not the state from
which the longer run resumes. The separately trained standard cell has one
cross-entropy objective. The two-pass cell uses equally weighted standard and shared
passes; both receive gradients. These objectives define different trained models.
The width-512 follow-up schedules in Appendix~\ref{app:consistency} instead use
approximately 2\% of their updates for warmup: 80 at 4,000 updates, 51 at 2,529,
and 155 at 7,766, with decay over the final 10\%.

\begin{table}[htbp]
\centering
\caption{Training endpoints. Token counts include the complete path to each endpoint;
extra branches consume additional experimental computation but do not add to that path.}
\label{tab:app-schedule}
\begin{tabular}{rrr}
\toprule
Endpoint updates & Decay starts & Training tokens \\
\midrule
4,000 & 3,600 & 1,048,576,000 \\
8,000 & 7,200 & 2,097,152,000 \\
16,000 & 14,400 & 4,194,304,000 \\
\bottomrule
\end{tabular}
\end{table}

The ladder and mechanism factorial initialize common tensors using generators keyed by
tensor name and training seed. Adding a gate or adapter therefore does not redraw the
shared backbone. The fixed-depth Huginn-style continuation also inherits a
named-initialized checkpoint; its recurrence and stochastic hidden-state initialization
still differ from the cell's. The three width-768 ladder seeds vary initialization
and training RNG while preserving data order. They replicate the two-pass-cell versus
untied comparison, not every ablation.

\section{Evaluation protocols and uncertainty}
\label{app:evaluation}

\paragraph{Chunk validation.}
Standard and parallel shared-cache validation each score 10,485,760 targets in the same
1,280 chunks of 8,192 tokens. Standard attention reads the cache for its own block and
iteration. In the main two-pass models' parallel shared pass, every core block reads older-token entries from
the last core slot's final iteration in the first pass; its current token uses its own
key and value. The two passes share the prologue computation but run separate core and
epilogue computations. Appendix~\ref{app:consistency} specifies the early-producer and
additional-sweep variants. Ground-truth token inputs are used throughout.

\paragraph{Prompt-prefill continuation.}
The primary continuation evaluation uses 1,280 documents, each at least 2,561 tokens
long. Each example starts at a document boundary and supplies 2,048 prompt tokens,
followed by 512 scored next-token targets: 655,360 targets in total. The full-cache
reference evaluates the ordinary model on the same document prefix and scores precisely
those targets. The shared-cache evaluation prefills the prompt with ordinary attention,
keeps the first pass's final donor and the prologue/epilogue caches, and then appends
new donor entries after each input token. The final prompt logit scores the first
continuation target; processing that target as the next ground-truth input produces the
prediction for the following target. Thus 512 targets require 511 new token updates
after prefill. No model samples its next token in this evaluation.

The diagnostic split uses a 512-token prompt and 2,048 targets on the same documents,
for 2,621,440 targets. It changes both prompt length and scored positions, so its
aggregate difference is not a controlled effect of continuation length alone. Positionwise
loss differences within either setting show how errors change as recursively generated
cache entries replace prompt entries. A second prefill diagnostic retains the first-pass
core donor but takes epilogue history and final prompt logits from the parallel second
pass. It still does not replace subsequent donor entries with first-pass entries.

\paragraph{Cold sequential evaluation.}
A separate diagnostic processes complete validation chunks token by token from empty
caches, with caches reset at document boundaries. For example, at width 768 and 4.19B
training tokens, the untreated two-pass model has cold shared-cache loss 3.3426 and
parallel shared-cache loss 3.2675. Their difference, 0.0751, describes that cold protocol;
it is not a correction to prompt-prefill continuation. Cold decoding and the prefill
protocols both use ground-truth inputs, despite their different cache histories.

\paragraph{Selection and confirmation.}
The original continuation documents lie beyond the chunk-validation prefix in the
validation shard. They were reused to compare objectives and regularization weights.
A further 1,280 eligible documents, disjoint from both earlier sets, were reserved for
confirmation of the selected width-768 donor-only recipe at $\gamma=0.3$. The two
training seeds were evaluated on this same confirmation set; they are not two new
independent text sets. The confirmation supports transfer to fresh documents from the
same source, not to another corpus. Cross-width parameter matching uses the original
spans and has no corresponding held-out confirmation.

All paired differences are computed before rounding from per-example loss sums on
identical targets. Unless stated otherwise, reported standard errors use 2,000
moving-block bootstrap resamples of eight adjacent chunks or spans. Blocks preserve
some dependence between nearby chunks; spans themselves are document-contained.
These errors condition on the trained checkpoint pair. They do not include training
seed variation or uncertainty from selecting a recipe on the reused set. Seed standard
deviations are reported separately where independent training seeds exist.

\section{Parameter, computation, and storage accounting}
\label{app:flops}

Table~\ref{tab:app-resources} counts all trainable parameters, including the tied
embedding once. Non-embedding counts exclude that table. All listed parameters
participate in the forward computation; shared tensors are counted once. The standard cell,
two-pass cell, and repaired cell at a given width have the same parameters.

For a dense matrix multiply, one multiply and one add count as two FLOPs. Attention
contributes $4d\bar{k}$ per block application, where for a window $W$ and chunk length
$L=8192$,
\begin{equation}
 \bar{k}=\frac{W(W+1)/2+(L-W)W}{L}.
\end{equation}
This averages the causal window over chunk positions and ignores additional document
boundaries. It is a reference accounting convention, not measured device work or an
exact cost for each continuation position. The counts include projection, gate,
adapter, MLP, and output-head matrix multiplications, but exclude normalization,
rotations, nonlinearities, masking, memory movement, and auxiliary-loss overhead.

Let $F_P$ denote the prologue cost and $F_Q$ the core, epilogue, and head together.
A standard forward costs $F_P+F_Q$; the two-pass objective costs $F_P+2F_Q$.
With full backpropagation and no activation recomputation, estimated training cost is
three times the executed forward. For truncated backpropagation, we count the executed
forward plus twice the taped forward; activation checkpointing adds the recomputed
core forward. Total estimated training FLOPs equal the appropriate per-token cost
times the number of training tokens. The fixed-depth repair uses the same two-pass
base accounting, with its auxiliary computations excluded explicitly.

\begin{table}[htbp]
\centering
\caption{Resources at three recurrent iterations. Parameters are millions; FLOPs are
millions per token. Forward is a single standard forward for every row; training counts
both passes for the two-pass objective. The repair shares the two-pass row's base cost.
$\Delta P$ and $\Delta F$ are percentage differences in total parameters and forward
FLOPs relative to the same-width untied control.}
\label{tab:app-resources}
\small
\setlength{\tabcolsep}{4pt}
\begin{tabular}{rlrrrrrr}
\toprule
Width & Trained model & \shortstack{Total\\params} & Non-emb. & $\Delta P$ (\%) & \shortstack{Forward\\FLOPs} & $\Delta F$ (\%) & \shortstack{Training\\FLOPs} \\
\midrule
384 & Untied & 43.66 & 24.34 & --- & 115.8 & --- & 347.4 \\
384 & Standard cell & 30.97 & 11.66 & -29.1 & 117.6 & +1.5 & 352.7 \\
384 & Two-pass cell & 30.97 & 11.66 & -29.1 & 117.6 & +1.5 & 687.5 \\
512 & Untied & 70.01 & 44.25 & --- & 178.0 & --- & 534.0 \\
512 & Standard cell & 46.93 & 21.18 & -33.0 & 181.1 & +1.8 & 543.4 \\
512 & Two-pass cell & 46.93 & 21.18 & -33.0 & 181.1 & +1.8 & 1058.2 \\
768 & Untied & 135.98 & 97.34 & --- & 328.9 & --- & 986.8 \\
768 & Standard cell & 85.24 & 46.61 & -37.3 & 336.0 & +2.2 & 1008.0 \\
768 & Two-pass cell & 85.24 & 46.61 & -37.3 & 336.0 & +2.2 & 1960.6 \\
1024 & Untied & 228.49 & 176.98 & --- & 532.9 & --- & 1598.8 \\
1024 & Standard cell & 136.20 & 84.69 & -40.4 & 545.5 & +2.4 & 1636.6 \\
1024 & Two-pass cell & 136.20 & 84.69 & -40.4 & 545.5 & +2.4 & 3180.4 \\
\bottomrule
\end{tabular}
\end{table}

The recurrent standard forward is 1.5--2.4\% more expensive than its same-width untied
control under this convention. Two-pass training instead costs 1.98--1.99 times as
much at the same token budget. Matching tokens therefore does not match training
computation. The repaired width-1024 versus untied width-768 comparison has almost
identical parameter counts ($+0.16\%$ for the recurrent model) but costs 1.66 times
as much per standard forward and
3.22 times as much in estimated training FLOPs. The existing intermediate endpoints
allow an approximately equal-training-compute comparison at fixed width: two-pass
models trained on 2.097B tokens versus untied models trained on 4.194B tokens
(Table~\ref{tab:app-training-compute}). The untied models receive at most 1.1\%
more training computation. The two-pass models have fewer parameters but
higher loss by 0.0957--0.1439 nats. These are parallel chunk losses before donor
repair; they do not measure shared-cache continuation or predict an unrun repair.

\begin{table}[htbp]
\centering
\caption{Approximately matched estimated training computation, using existing
seed-0 endpoints. T denotes the two-pass model at 2.097B tokens and U the untied
model at 4.194B. Parameters include embeddings; computation is total training
FLOPs in units of $10^{18}$. Losses score the same 1,280 chunks (10,485,760 targets):
T uses parallel shared-cache validation before repair, U standard validation.
Parentheses are paired block-bootstrap SEs, conditional on the checkpoints.}
\label{tab:app-training-compute}
\small
\setlength{\tabcolsep}{4pt}
\begin{tabular}{rrrrrrr}
\toprule
Width & \shortstack{Params (M)\\T / U} & \shortstack{Training FLOPs\\T / U} & \shortstack{Compute\\ratio T/U} & \shortstack{T loss\\$\downarrow$} & \shortstack{U loss\\$\downarrow$} & $\Delta$ loss T$-$U \\
\midrule
384 & 30.97 / 43.66 & 1.442 / 1.457 & 0.989 & 3.6257 & 3.4819 & $+0.1439\ (0.0006)$ \\
512 & 46.93 / 70.01 & 2.219 / 2.240 & 0.991 & 3.4967 & 3.3710 & $+0.1257\ (0.0006)$ \\
768 & 85.24 / 135.98 & 4.112 / 4.139 & 0.993 & 3.3448 & 3.2394 & $+0.1054\ (0.0008)$ \\
1024 & 136.20 / 228.49 & 6.670 / 6.706 & 0.995 & 3.2514 & 3.1557 & $+0.0957\ (0.0008)$ \\
\bottomrule
\end{tabular}
\end{table}

\paragraph{Persistent cache storage.}
For $S$ cache sets, BF16 storage per cached token is
$2\times2\times H_{\mathrm{KV}}\times64\times S$ bytes. At P1E2, $K=4$, and $R=3$,
standard caching needs $1+4\cdot3+2=15$ sets; full sharing needs $1+1+2=4$.
Uniform 2,048-position rings therefore reduce allocated KV storage by $15/4=3.75$.
Sizing each set to its longest read window instead gives five long and ten short
sets for standard caching, versus three long and one short set for full sharing:
$(5\cdot2+10)/(3\cdot2+1)=20/7\simeq2.86$.
The shared donor must retain a long window because the first core slot reads it,
although the last core slot produces it. These formulas exclude scratch tensors
and model weights; measured peaks appear in Appendix~\ref{app:hardware}.

\section{Complete width and token ladder}
\label{app:ladder}
\label{app:ablations}

\begin{figure}[htbp]
  \centering
  \includegraphics[width=\linewidth]{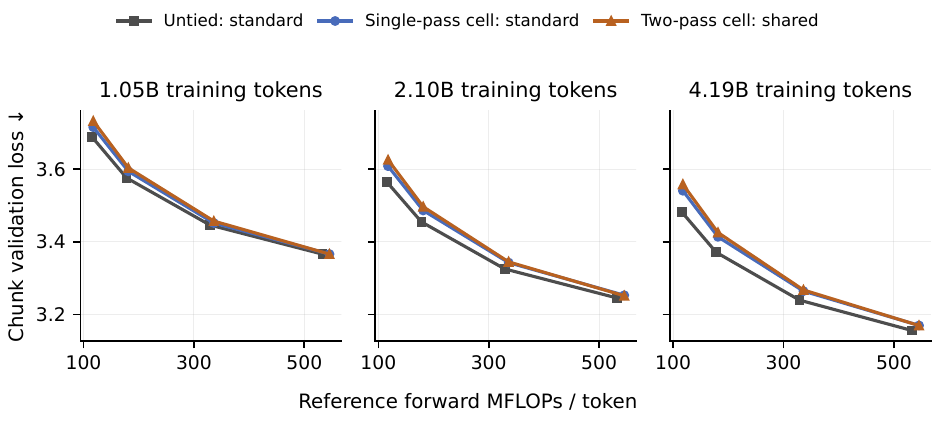}
  \caption{\textbf{Loss versus reference forward computation across training-token budgets.}
  The same checkpoints and losses as Figure~\ref{fig:ladder}, with widths 384, 512,
  768, and 1024 in increasing compute order. The horizontal axis counts one standard
  forward under Appendix~\ref{app:flops}'s attention-length convention, as an inference
  compute reference; it does not count two-pass training or measure serving latency.
  Untied and single-pass cell losses use standard validation; two-pass cell losses
  use parallel shared-cache validation before consistency repair. All points use
  training seed 0 and the same 10,485,760 targets.}
  \label{fig:app-ladder-flops}
\end{figure}

Table~\ref{tab:app-ladder} separates the ordinary cell trained with one loss from
the cell trained with the two-pass objective. The latter has two reported parallel
losses, neither of which is a recursively updated shared-cache continuation loss.
These are seed-0 results. Increasing training tokens improves every arm in this
range, while both recurrent-minus-untied gaps widen at each tested width.
Figure~\ref{fig:app-ladder-flops} shows the same ladder against reference forward
FLOPs: the gaps widen while the same-width compute mismatch remains 1.5--2.4\%.
These measurements
do not establish an asymptotic capacity floor or a width at which parity must occur.

\begin{table}[htbp]
\centering
\caption{Complete chunk-validation ladder; loss $\downarrow$. Tokens are billions.
$\Delta_{\rm cell}$ is standard-cell minus untied; $\Delta_{\rm shared}$ is two-pass
shared minus untied. Each row scores the same 10,485,760 targets.}
\label{tab:app-ladder}
\small
\setlength{\tabcolsep}{5pt}
\begin{tabular}{rrrrrrrr}
\toprule
Width & Tokens & Untied & Std. cell & Two-pass std. & Two-pass shared & $\Delta_{\rm cell}$ & $\Delta_{\rm shared}$ \\
\midrule
384 & 1.049 & 3.6880 & 3.7158 & 3.7338 & 3.7329 & +0.0278 & +0.0449 \\
384 & 2.097 & 3.5641 & 3.6083 & 3.6271 & 3.6257 & +0.0442 & +0.0617 \\
384 & 4.194 & 3.4819 & 3.5408 & 3.5594 & 3.5582 & +0.0589 & +0.0763 \\
512 & 1.049 & 3.5759 & 3.5941 & 3.6076 & 3.6035 & +0.0182 & +0.0276 \\
512 & 2.097 & 3.4549 & 3.4864 & 3.5011 & 3.4967 & +0.0315 & +0.0417 \\
512 & 4.194 & 3.3710 & 3.4138 & 3.4292 & 3.4256 & +0.0428 & +0.0546 \\
768 & 1.049 & 3.4462 & 3.4517 & 3.4624 & 3.4571 & +0.0055 & +0.0109 \\
768 & 2.097 & 3.3254 & 3.3427 & 3.3504 & 3.3448 & +0.0173 & +0.0194 \\
768 & 4.194 & 3.2394 & 3.2643 & 3.2728 & 3.2675 & +0.0249 & +0.0281 \\
1024 & 1.049 & 3.3661 & 3.3662 & 3.3720 & 3.3662 & +0.0002 & +0.0001 \\
1024 & 2.097 & 3.2451 & 3.2529 & 3.2577 & 3.2514 & +0.0077 & +0.0063 \\
1024 & 4.194 & 3.1557 & 3.1697 & 3.1758 & 3.1694 & +0.0140 & +0.0137 \\
\bottomrule
\end{tabular}
\end{table}

At 4.19B tokens, sharing reduces the two-pass model's own parallel loss by
0.0012, 0.0036, 0.0053, and 0.0063 at widths 384, 512, 768, and 1024, respectively.
This within-model contrast should not be substituted for the cost of changing the
training objective or the cost of recursive cache updates. The paired uncertainty
of the separately trained standard-cell versus untied comparison is approximately
0.0004--0.0005 nats; that of the two-pass shared versus untied comparison is
approximately 0.0004--0.0006 at this endpoint.

\section{Training seeds and held-out confirmation}
\label{app:seeds}

\begin{table}[htbp]
\centering
\caption{Three training seeds at width 768, on the common chunk evaluation. Loss
$\downarrow$; $\Delta$ is two-pass shared minus untied. Tokens are billions.}
\label{tab:app-seeds}
\begin{tabular}{rrrrrr}
\toprule
Tokens & Seed & Untied & Two-pass std. & Two-pass shared & $\Delta$ \\
\midrule
1.049 & 0 & 3.4462 & 3.4624 & 3.4571 & +0.0109 \\
1.049 & 1 & 3.4476 & 3.4637 & 3.4587 & +0.0111 \\
1.049 & 2 & 3.4444 & 3.4620 & 3.4569 & +0.0125 \\
2.097 & 0 & 3.3254 & 3.3504 & 3.3448 & +0.0194 \\
2.097 & 1 & 3.3264 & 3.3531 & 3.3479 & +0.0215 \\
2.097 & 2 & 3.3235 & 3.3512 & 3.3455 & +0.0220 \\
4.194 & 0 & 3.2394 & 3.2728 & 3.2675 & +0.0281 \\
4.194 & 1 & 3.2372 & 3.2753 & 3.2702 & +0.0330 \\
4.194 & 2 & 3.2368 & 3.2747 & 3.2685 & +0.0317 \\
\bottomrule
\end{tabular}
\end{table}

The mean shared-minus-untied differences at 1.05B, 2.10B, and 4.19B tokens are
$0.0115$, $0.0209$, and $0.0309$, with seed standard deviations
$0.0009$, $0.0014$, and $0.0025$ ($n=3$). These are larger than the paired text-sampling
errors and specific to this comparison. Across the same three seeds at 4.19B,
shared-minus-standard loss within the two-pass model averages $-0.0055$ with seed
standard deviation $0.0005$.

\begin{table}[htbp]
\centering
\caption{Held-out width-768 confirmation at 4.19B tokens, with 2,048 prompt tokens
and 512 targets. Differences compare the repair to its own untreated parent; parentheses
are paired bootstrap SEs. All losses are in nats per target ($\downarrow$).}
\label{tab:app-heldout}
\small
\begin{tabular}{rrrrrr}
\toprule
Seed & Parent shared & Repaired shared & Shared difference & Full-cache difference & Untied \\
\midrule
0 & 3.3346 & 3.2911 & $-0.0434\ (0.0011)$ & $+0.0010\ (0.0002)$ & 3.2390 \\
1 & 3.3235 & 3.2936 & $-0.0298\ (0.0006)$ & $+0.0014\ (0.0003)$ & 3.2336 \\
\bottomrule
\end{tabular}
\end{table}

On these fresh documents, the repair reduces the within-model continuation penalty
from 0.0554 to 0.0110 for seed 0 and from 0.0416 to 0.0104 for seed 1: reductions of
80\% and 75\%. Remaining repaired-minus-untied gaps are $0.0521\ (0.0015)$ and
$0.0600\ (0.0017)$. This supports the repair across two trained checkpoints while
retaining both its small full-cache cost and its residual gap to the untied model.

\section{Consistency objective and ablations}
\label{app:consistency}
\label{app:cachedetail}

For a donor tensor $X\in\{K,V\}$, the auxiliary error is
\begin{equation}
 e_X=\frac{\operatorname{mean}\bigl[(X_2-\operatorname{stopgrad}(X_1))^2\bigr]}
 {s_X+10^{-6}},\qquad L_{\rm donor}=\tfrac12(e_K+e_V).
\end{equation}
The target is the first pass's final-iteration own key or value in the last core slot;
the prediction is the corresponding tensor in the second pass. Keys are compared
after query/key normalization and positional rotation, values as stored. For donor-only
training, the objective is
$L=\tfrac12\mathrm{CE}_1+\tfrac12\mathrm{CE}_2+\gamma L_{\rm donor}$.
Only the auxiliary reference tensor detaches: gradients through the first-pass donor
used as attention input to the second pass remain live.

The fixed scales $s_K,s_V$ are reference mean squares, calibrated once at branch start.
Each rank evaluates one microbatch of two 8,192-token chunks; averaging across eight
ranks uses 16 chunks, or 131,072 tokens. These are the first microbatches on each rank
of the branch's starting update, not necessarily 16 consecutive chunks in the global
stream. Calibration runs at depth 3 for fixed-depth models and at depth 6 for the
variable-depth model. Scales are held fixed thereafter and stored in checkpoints.
The initial calibration sample depends on rank count and microbatch size; changing
hardware decomposition requires preserving the scales or treating recalibration as
a recipe change.

For epilogue supervision, K/V errors are averaged within each epilogue block and then
across the two blocks. The combined objective uses
$L_{\rm KV}=\tfrac12(L_{\rm donor}+L_{\rm epi})$, whereas donor-only uses
$L_{\rm KV}=L_{\rm donor}$. At equal $\gamma$, dropping the epilogue term therefore
doubles the donor coefficient. The comparison establishes that donor-only supervision
is sufficient for the observed repair; it does not isolate the effect of removing
an otherwise equally weighted epilogue penalty.

Output distillation \citep{hinton2015distilling} uses
$\beta\,\operatorname{mean}_{t}\mathrm{KL}(\operatorname{stopgrad}(p_{1,t})\Vert p_{2,t})$
at temperature one. Matching output distributions need not match stored cache entries,
and matching entries in the parallel objective does not guarantee a fixed point under
recursive decoding. The continuation evaluations test the consequences directly.

\begin{table}[htbp]
\centering
\caption{Width-768 auxiliary-loss ablations on the same 1,600-update decay branch.
Full and shared losses score the primary continuation targets ($\downarrow$).
Penalty is shared minus full within that model. The last column is the change in
\emph{chunk-standard} loss relative to the no-auxiliary branch control; its preset
quality limit is $+0.005$.}
\label{tab:app-consistency}
\small
\setlength{\tabcolsep}{4pt}
\begin{tabular}{lrrrr}
\toprule
Auxiliary objective & Full & Shared & Penalty & $\Delta$ chunk std. \\
\midrule
None & 3.2827 & 3.3353 & +0.0526 & +0.0000 \\
Donor + epilogue, $\gamma=0.1$ & 3.2826 & 3.3029 & +0.0203 & +0.0001 \\
Donor + epilogue, $\gamma=0.3$ & 3.2834 & 3.2956 & +0.0122 & +0.0010 \\
Donor only, $\gamma=0.1$ & 3.2832 & 3.3001 & +0.0169 & +0.0004 \\
Donor only, $\gamma=0.3$ & 3.2839 & 3.2938 & +0.0099 & +0.0013 \\
Donor only, $\gamma=1.0$ & 3.2884 & 3.2902 & +0.0018 & +0.0052 \\
Epilogue only, $\gamma=0.1$ & 3.2826 & 3.3251 & +0.0425 & +0.0001 \\
Donor + epilogue, $\gamma=0.1$, $\beta=1$ & 3.2938 & 3.2927 & -0.0012 & +0.0112 \\
Donor only, $\gamma=0.1$, $\beta=0.3$ & 3.2852 & 3.2926 & +0.0075 & +0.0029 \\
\bottomrule
\end{tabular}
\end{table}

The control branch reproduces the parent's chunk-standard loss to 0.0001 and its
primary continuation loss to 0.0006. Donor-only $\gamma=0.3$ provides a substantial
improvement while staying inside the quality limit. Increasing $\gamma$ to 1.0 further
reduces the residual penalty but raises chunk-standard loss by 0.0052 and primary
full-cache continuation loss by 0.0057. The combined K/V and $\beta=1$ distillation
arm almost closes the within-model penalty, at a chunk-standard cost of 0.0112.
This is a result for that combined objective, not an isolated estimate of the KL
term under the decay schedule. A smaller $\beta=0.3$ with donor $\gamma=0.1$ gives
similar served loss to donor-only $\gamma=0.3$, with a larger full-cache cost.
The chosen coefficient is a tested tradeoff, not an estimated optimum.

\paragraph{Repair with different recurrent updates.}
\label{app:shared-update-controls}
A width-512 pilot applies the donor-only term to the gated cell, plain normalized
overwrite and Huginn-style injection. Each update has one 3,600-step trunk followed
by paired 400-step branches with $\gamma=0$ or $0.3$. The injection uses Gaussian
latents keyed by example and position, shared across parallel passes and sequential
decoding, rather than the earlier stream-noise convention.
Table~\ref{tab:app-shared-update-controls} scores all six models under cold serving
on the same 1,280 selection documents, distinct from the main evaluation and the
producer confirmation below. Repair improves all three updates in this single-seed
pilot; the effect is not specific to the gated carry. The paired errors quantify
variation over documents, not training seeds.

\begin{table}[htbp]
\centering
\caption{Recurrent-update repair pilot: width 512, P1E2, $R=3$, 1.048576B training
tokens and four persistent cache sets (4\,KiB/token). Cold loss scores 512 targets
after 2,048 prompt tokens per document. $\Delta$ is repair minus the corresponding
untreated branch, with paired bootstrap SE. Reference forward cost follows
Appendix~\ref{app:flops}.}
\label{tab:app-shared-update-controls}
\small
\setlength{\tabcolsep}{4pt}
\begin{tabular}{lrrrrr}
\toprule
Update & $\gamma$ & Params (M) & Fwd MFLOPs/tok & Cold loss $\downarrow$ & $\Delta$ loss $\downarrow$ (SE) \\
\midrule
Cell & 0 & 46.933 & 181.1 & 3.7961 & --- \\
Cell & 0.3 & 46.933 & 181.1 & 3.6587 & $-0.1374\;(0.0026)$ \\
Plain & 0 & 46.408 & 178.0 & 3.7658 & --- \\
Plain & 0.3 & 46.408 & 178.0 & 3.6844 & $-0.0814\;(0.0013)$ \\
Injection & 0 & 46.932 & 181.1 & 3.6945 & --- \\
Injection & 0.3 & 46.932 & 181.1 & 3.6674 & $-0.0271\;(0.0005)$ \\
\bottomrule
\end{tabular}
\end{table}

\paragraph{Repeated donor updates over whole chunks.}
\label{app:donor-iterate}
A smaller diagnostic repeatedly applies the parallel cache producer to the same
64 validation chunks, starting from standard attention. Let $k$ count whole-sequence
forwards, each with three core iterations. At $k=1$, the model produces standard
attention caches; at $k=2$, it reads those donors, as in the two-pass objective.
For each $k>2$, it restarts from the same prologue states and reads the preceding
forward's donors, then recomputes its own donor and epilogue outputs.
Table~\ref{tab:app-donor-iterate} reports the loss of each forward on the same targets.

\begin{table}[htbp]
\centering
\caption{Recorded donor-iteration diagnostic at width 768 and 4.194B training
tokens, seed 0: 64 chunks of 8,192 targets (524,288 total). Each $k$ is a new
whole-sequence forward with three core iterations, not an additional core iteration
within one forward. Loss $\downarrow$; the first forward uses standard attention.}
\label{tab:app-donor-iterate}
\begin{tabular}{lrrrrrr}
\toprule
Checkpoint & $k=1$ & $k=2$ & $k=3$ & $k=4$ & $k=5$ & $k=6$ \\
\midrule
Untreated & 3.2568 & 3.2510 & 3.2994 & 3.3246 & 3.3290 & 3.3313 \\
Late donor consistency & 3.2579 & 3.2515 & 3.2596 & 3.2640 & 3.2663 & 3.2671 \\
\bottomrule
\end{tabular}
\end{table}

The second forward improves loss in both checkpoints, but subsequent donor updates
increase it. Late consistency substantially reduces that increase. No standard-prefill
prefix is retained during these whole-chunk updates: this is a cold-history diagnostic,
not a decomposition of the primary prompt-prefill continuation penalty.
Six iterates do not establish convergence or equality to sequential decoding.

\paragraph{An early cache producer.}
\label{app:early-producer}
An alternative forms each token's shared core K/V directly from its prologue
representation, using the first core block's existing projections. This entry is
independent of the recurrent computation, so parallel shared attention and cold
sequential serving use the same past-token entries. A single shared-pass training
objective suffices; a standard producer pass is unnecessary. Producer depth is also
a design choice in cross-layer and decoder-sharing architectures
\citep{brandon2024cla,sun2024yoco,wu2025systematic}.

This removes a path for communicating refined representations between tokens.
Later recurrent passes can change the current token's queries, carry and prediction,
but cannot change the entry read by future core blocks. Their output still reaches
the epilogue, whose separate caches retain projections of recurrently refined states
for future epilogue attention. Thus recurrence can improve computation over a fixed
memory; it cannot enrich that memory for future recurrent computation. The early
producer and its prologue remain trainable end to end.

Table~\ref{tab:app-early-producer} compares early and repaired late producers at
width 512, with the same P1E2 layout and $R=3$. The early models use either the
two-pass objective or only shared-pass cross-entropy, without consistency. All
losses use cold serving on the same 1,280 selection documents and 655,360 targets
as the refinement ablation below. This avoids applying the single-objective model's
untrained standard-attention function to the prompt.

\begin{table}[htbp]
\centering
\caption{Early versus repaired late cache production. All models have 46.933M
parameters and four persistent cache sets (4\,KiB per token in BF16).
Training cost is relative to the repaired late model. Reference forward FLOPs
use Appendix~\ref{app:flops}'s convention plus the early producer's extra K/V
projection; they are not measured latency or prefill cost. Cold loss $\downarrow$.}
\label{tab:app-early-producer}
\small
\setlength{\tabcolsep}{4pt}
\begin{tabular}{llrrrr}
\toprule
Producer & Training objective & Tokens (B) & Train cost & Fwd MFLOPs/tok & Cold loss \\
\midrule
Late & Two-pass + consistency & 1.049 & $1.000\times$ & 181.1 & 3.6587 \\
Early & Two-pass & 1.049 & $1.001\times$ & 181.7 & 3.7362 \\
Early & Shared CE & 1.049 & $0.515\times$ & 181.7 & 3.7247 \\
Early & Shared CE & 2.036 & $1.000\times$ & 181.7 & 3.6240 \\
\bottomrule
\end{tabular}
\end{table}

At equal tokens, the repaired late producer has lower cold loss than either early
variant. The single-objective early producer costs about half as much to train per
token; spending the same training compute on more tokens reverses the comparison
(3.6240 versus 3.6587). These are single-seed recipe comparisons, which change both
cache production and training supervision; they do not isolate the effect of
storing deeper features.

We then compare two existing checkpoints on 1,280 previously unused documents
under the same cold protocol (Table~\ref{tab:app-producer-closeout}). The late producer
trained for 4.194B tokens has lower loss than the early producer trained for 2.036B:
early minus late is $+0.1409$ (paired SE $0.0019$; 95\% bootstrap interval
$[0.1374,0.1448]$). Both have nearly the same reference inference cost and cache
storage. The late model used $2.06\times$ as many tokens and approximately
$4\times$ the estimated training FLOPs, with a different objective and schedule.
This establishes an ordering of these available checkpoints at similar inference
cost, not a general ranking of producer architectures or training efficiency.

\begin{table}[htbp]
\centering
\caption{Producer comparison on fresh documents, separate from the pilot selection
set. Both width-512 P1E2 checkpoints have 46.933M parameters, $R=3$ and four cache
sets (4\,KiB/token). Cold loss scores 655,360 targets with 2,048-token prompts.
Training FLOPs exclude auxiliary-loss overhead; forward FLOPs are reference
estimates, not measured serving latency.}
\label{tab:app-producer-closeout}
\small
\setlength{\tabcolsep}{5pt}
\begin{tabular}{lrrrr}
\toprule
Producer & Tokens (B) & Train FLOPs ($10^{18}$) & Fwd MFLOPs/tok & Cold loss $\downarrow$ \\
\midrule
Late, two-pass + consistency & 4.194 & 4.4382 & 181.144 & 3.4494 \\
Early, shared CE & 2.036 & 1.1095 & 181.668 & 3.5903 \\
\bottomrule
\end{tabular}
\end{table}

The early producer still uses its recurrent depth. In a 64-chunk monitoring probe
of the longer-trained early model, parallel shared losses at depths $2,3,4,6$ are
$3.7838,3.5717,3.6362,4.0274$, respectively (524,288 targets; trained depth three).
Removing the third pass hurts this checkpoint, as do passes beyond the trained depth.
This is a depth perturbation of one model, not a comparison of separately trained
depths or a held-out confirmation of the best architecture.

\paragraph{Training with additional cache-refinement sweeps.}
\label{app:refinement-training}
We also trained width-512 P1E2 models with four distinct core blocks and two or
four whole-sequence sweeps, keeping the recurrent depth at $R=3$ (46.933M parameters).
Each sweep restarts the recurrent states
from the same prologue representations. Sweep 1 uses standard attention;
each later sweep reads the preceding sweep's donor caches. The objective is
$\tfrac12\mathrm{CE}_1+\tfrac12\mathrm{CE}_n$ for $n\in\{2,4\}$, with gradients
through all sweeps. Intermediate sweeps execute only the core; the epilogue and
vocabulary head run on sweeps 1 and $n$. Treated branches add donor-only consistency with $\gamma=0.3$
over the final 10\% of updates, matching the last sweep to the detached first-sweep
reference with fixed calibrated loss scales. This varies the sweep count within our recipe; it does not reproduce
the full LCKV training algorithm \citep{wu2024lckv}.

Table~\ref{tab:app-refinement-training} separates equal-token and approximately
equal-training-compute comparisons. All use one training seed and the same new
selection set of 1,280 documents, with 2,048 prompt tokens and 512 continuation
targets per document (655,360 targets). The set is separate from the earlier
selection and confirmation documents. Handoff serving retains standard-prefill
prompt caches; cold serving processes the prompt and continuation under the shared
policy from empty caches. Both score identical ground-truth targets.

\begin{table}[htbp]
\centering
\caption{Cache-refinement training ablation at width 512. Training compute is
relative to the two-sweep, 1.048576B-token reference (1,109.560 estimated PFLOPs),
using the accounting in Appendix~\ref{app:flops}.
Four-sweep compute-matched runs use 2,529 updates (a $+0.0183\%$ compute mismatch);
the other runs use 4,000. All sequential evaluations retain $R=3$ and one shared
core cache; they do not execute additional sweeps. Loss $\downarrow$.}
\label{tab:app-refinement-training}
\small
\setlength{\tabcolsep}{5pt}
\begin{tabular}{rrrrrrr}
\toprule
Sweeps $n$ & $\gamma$ & Tokens (B) & Train compute & Handoff & Cold & Chunk std. \\
\midrule
2 & 0 & 1.049 & $1.00\times$ & 3.7122 & 3.7961 & 3.6081 \\
2 & 0.3 & 1.049 & $1.00\times$ & 3.6523 & 3.6587 & 3.6101 \\
4 & 0 & 1.049 & $1.58\times$ & 3.6684 & 3.6522 & 3.6100 \\
4 & 0.3 & 1.049 & $1.58\times$ & 3.6733 & 3.6539 & 3.6113 \\
\midrule
4 & 0 & 0.663 & $1.00\times$ & 3.7592 & 3.7484 & 3.7173 \\
4 & 0.3 & 0.663 & $1.00\times$ & 3.7632 & 3.7515 & 3.7187 \\
\bottomrule
\end{tabular}
\end{table}

At equal tokens, four sweeps without consistency improve cold loss by 0.0065
relative to the repaired two-sweep model (paired bootstrap SE 0.0011), but worsen
handoff loss by 0.0161 and require 58\% more training compute. At matched compute,
the four-sweep model sees fewer tokens and is worse under both serving protocols;
its cold loss is 0.0898 higher. Adding consistency to four-sweep training improves
neither protocol in this pilot. Thus four sweeps do not improve the served-loss
tradeoff at the common training budget. This is a negative result for the tested
four-sweep recipe at matched compute; the equal-token cold improvement remains
a benefit at higher training cost. These results use one seed and selection data.

\begin{table}[htbp]
\centering
\caption{All six donor-refinement readouts for the four-sweep-trained models,
on 64 chunks of 8,192 targets (524,288 total). The trained shared objective uses
sweep 4. Cold sequential loss scores the same chunks. Loss $\downarrow$;
these chunk losses are separate from the continuation losses above.}
\label{tab:app-refinement-readouts}
\small
\setlength{\tabcolsep}{4pt}
\begin{tabular}{lrrrrrrr}
\toprule
Tokens, $\gamma$ & $k=1$ & $k=2$ & $k=3$ & $k=4$ & $k=5$ & $k=6$ & Cold \\
\midrule
1.049B, 0 & 3.6001 & 3.6525 & 3.6482 & 3.5882 & 3.6413 & 3.6215 & 3.6073 \\
1.049B, 0.3 & 3.6016 & 3.6639 & 3.6501 & 3.5886 & 3.6526 & 3.6376 & 3.6099 \\
0.663B, 0 & 3.7170 & 3.7546 & 3.7462 & 3.7053 & 3.7328 & 3.7249 & 3.7175 \\
0.663B, 0.3 & 3.7185 & 3.7595 & 3.7523 & 3.7059 & 3.7475 & 3.7388 & 3.7213 \\
\bottomrule
\end{tabular}
\end{table}

The diagnostic in Table~\ref{tab:app-refinement-readouts} has its lowest loss at
the trained fourth sweep, followed by a rebound at sweep 5. It does not show
monotone convergence to sequential serving. The shared validation loss of these
models uses four sweeps, whereas the original models use two, so their parallel
losses also have different evaluation costs. The pattern is consistent with
specialization to the training sweep count, rather than evidence that extra sweeps
eliminate the recursive-cache discrepancy.

\paragraph{A restart schedule that did not preserve parent quality.}
An additional four-arm pilot began from the already annealed width-768 model and reset
the optimizer. It added 1,000 updates (262.144M tokens), with 80 warmup updates and decay
only over the final 100. Table~\ref{tab:app-rewarm} reports this separately because
even the no-auxiliary control degraded sharply. The cache term helps relative to that
control, but none of the resulting models improves upon the original parent's shared
continuation loss. The main repair instead resumes the original optimizer and learning-rate
schedule at the start of its decay segment.

\begin{table}[htbp]
\centering
\caption{Optimizer-reset pilot: primary continuation losses and separate chunk-standard
loss ($\downarrow$). The parent has 4.194B training tokens; pilot arms have 4.456B.}
\label{tab:app-rewarm}
\begin{tabular}{lrrr}
\toprule
Arm & Full-cache continuation & Shared continuation & Chunk standard \\
\midrule
Parent & 3.2827 & 3.3347 & 3.2728 \\
Reset control & 3.4093 & 3.4602 & 3.3917 \\
KL only, $\beta=1$ & 3.4207 & 3.4272 & 3.4035 \\
K/V, $\gamma=0.1$ & 3.4099 & 3.4287 & 3.3929 \\
Both & 3.4220 & 3.4199 & 3.4045 \\
\bottomrule
\end{tabular}
\end{table}

\paragraph{Applying consistency throughout training.}
A fresh fixed-depth pair at 1.049B tokens uses donor $\gamma=0.3$ from the start.
The reference scales track first-pass mean squares with an exponential moving average
(decay 0.99), updated after each step; each update uses fixed denominators across its
microbatches and ranks. Shared continuation loss improves from 3.5097 to 3.4839,
while chunk-standard loss increases from 3.4628 to 3.4654. A variable-depth pair at
2.097B improves shared continuation by 0.029--0.033 but increases full-cache
continuation loss by 0.006--0.009 (Appendix~\ref{app:adaptive}). These comparisons
change both the duration of regularization and its scale normalization relative to
late fixed-scale repair. They do not identify which change explains the larger cost.
The recorded whole-run models also include one warmup update to the EMA before the
training counter starts; this initialization detail is part of their evaluated recipe.

\section{Recurrent update and deep-depth controls}
\label{app:depth}

The fixed-depth factorial changes two components: gated carry accumulation versus a
plain normalized update, and presence versus absence of the prologue adapter.
It does not independently toggle gates while preserving the carry. At width 768,
three iterations, and 1.049B tokens, losses are 3.4517/3.4518 for the cell without/with
the adapter and 3.4838/3.4683 for the plain update. Averaging over adapter levels,
the gated-carry update improves loss by $0.0243\ (0.0002)$; the adapter improves it
by $0.0077\ (0.0002)$, with interaction $0.0156\ (0.0004)$. Each added gate or
adapter contributes approximately 1.18M matrix parameters. These are single-seed,
common-backbone controls, not parameter-identical treatments.

Table~\ref{tab:app-update-budget} collects all available training-budget endpoints
for the five updates in Table~\ref{tab:arms}, with the untied control for reference.
Among the recurrent updates, only the cell without an adapter and the Huginn-style
baseline were extended to 2.097B and 4.194B tokens. The Huginn-style model's gap to the cell stays close to
0.023 nats across these budgets. The longer-training behavior of the other three
ablations remains untested. This fixed-depth comparison differs from the deep
experiment below in both width and training-depth distribution.

\begin{table}[htbp]
\centering
\caption{\textbf{Update-rule comparisons across training-token budgets.}
Width 768, with three recurrent iterations for P1E2 and a 15-layer untied control.
Columns give standard validation loss ($\downarrow$) after the final learning-rate
decay at each budget, on the same 1,280 chunks (10,485,760 targets).
One training seed per arm; Huginn-style losses use the same evaluation-noise draw
as Table~\ref{tab:arms}. A dash in a loss column denotes an experiment not run at that budget.
Parentheses in the final row are paired block-bootstrap SEs, conditional on the
checkpoints.}
\label{tab:app-update-budget}
\begin{tabular}{llrrr}
\toprule
Update & Adapter & 1.049B tokens & 2.097B tokens & 4.194B tokens \\
\midrule
Gated carry & No & 3.4517 & 3.3427 & 3.2643 \\
Plain overwrite & No & 3.4838 & --- & --- \\
Gated carry & Yes & 3.4518 & --- & --- \\
Plain overwrite & Yes & 3.4683 & --- & --- \\
Huginn-style & Yes & 3.4745 & 3.3649 & 3.2891 \\
\midrule
Untied & --- & 3.4462 & 3.3254 & 3.2394 \\
\midrule
\multicolumn{2}{l}{Huginn-style minus gated carry (no adapter)} &
$+0.0228$ & $+0.0222$ & $+0.0248$ \\
\multicolumn{2}{l}{\quad (paired SE)} & (0.0003) & (0.0003) & (0.0003) \\
\bottomrule
\end{tabular}
\end{table}

For the deep comparison, all four width-512 models train on 1.049B tokens, with depth
drawn uniformly from 4 through 32 and synchronized across ranks. Full-gradient arms
checkpoint core activations and recompute them in the backward pass. Truncated arms
backpropagate through only the final eight iterations. Without an adapter, truncation
cuts the cell's gradient to its prologue when the sampled depth exceeds eight; the
injection model retains a path through its prologue adapter. The comparison therefore
tests an interaction between update rule and gradient path, rather than a generic
benefit of truncation.

\begin{table}[htbp]
\centering
\caption{Forced-depth standard losses ($\downarrow$) after uniform-depth training
at width 512. Each entry scores 10,485,760 targets; one training seed per arm.}
\label{tab:app-deep}
\begin{tabular}{llrrrrr}
\toprule
Update & Backpropagation & $R=4$ & $R=8$ & $R=16$ & $R=32$ & $R=64$ \\
\midrule
Cell & Full & 3.6656 & 3.6112 & 3.6028 & 3.6065 & 3.6287 \\
Cell & Last eight & 3.7076 & 3.6748 & 3.6731 & 3.6822 & 3.7107 \\
Huginn-style & Full & 3.6581 & 3.6052 & 3.6036 & 3.6036 & 3.6037 \\
Huginn-style & Last eight & 3.6472 & 3.5913 & 3.5896 & 3.5896 & 3.5896 \\
\bottomrule
\end{tabular}
\end{table}

At $R=32$, truncation increases cell loss by $0.0757\ (0.0004)$ and decreases
Huginn-style loss by $0.0140\ (0.0002)$. With full gradients, Huginn-style minus cell
is $-0.0029\ (0.0003)$; with truncation it is $-0.0926\ (0.0004)$.
The full-gradient losses are close in absolute terms, without establishing statistical
equivalence or reproducing the original Huginn scaling result. Gaussian latent seeds
were controlled across depths, and two evaluation-noise draws changed the aggregate
Huginn-style losses by at most 0.00012. These draws are not additional training seeds.

\begin{figure}[htbp]
\centering
\includegraphics[width=0.84\linewidth]{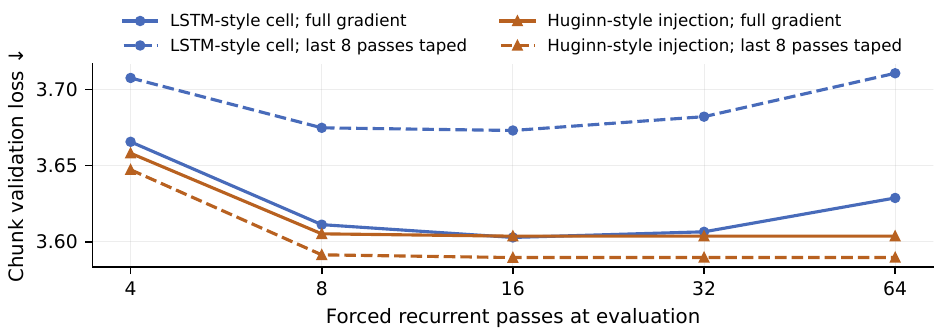}
\caption{Depth extrapolation with full and truncated gradients, using the models in
Table~\ref{tab:app-deep}. Training depths range from 4 to 32; depth 64 is extrapolation.}
\label{fig:app-depth}
\end{figure}

\section{Variable depth and stopping}
\label{app:adaptive}
\label{app:adaptivedetail}

The variable-depth two-pass model samples $R$ uniformly from 3 through 6, with full
backpropagation, and trains to 2.097B tokens. Table~\ref{tab:app-depth-robustness}
compares it with the two-pass model trained at fixed $R=3$. Variable-depth training
sacrifices some shortest-depth quality and remains much more tolerant of additional
iterations. This concerns parallel evaluation, not the recursively updated cache.

\begin{table}[htbp]
\centering
\caption{Parallel losses at width 768 and 2.097B tokens ($\downarrow$). A dash denotes
an unrecorded fixed-depth comparison.}
\label{tab:app-depth-robustness}
\begin{tabular}{rrrrr}
\toprule
Evaluation $R$ & \shortstack{Fixed $R=3$\\standard} & \shortstack{Fixed $R=3$\\shared} & \shortstack{Variable $R$\\standard} & \shortstack{Variable $R$\\shared} \\
\midrule
3 & 3.3504 & 3.3448 & 3.3693 & 3.3577 \\
4 & 3.4142 & 3.4691 & 3.3513 & 3.3452 \\
5 & --- & --- & 3.3477 & 3.3436 \\
6 & 3.7543 & 3.9704 & 3.3483 & 3.3449 \\
8 & --- & --- & 3.3551 & 3.3515 \\
\bottomrule
\end{tabular}
\end{table}

The late adaptive repair starts from update 7,000 and repeats the final 1,000 updates
with donor-only $\gamma=0.3$. Because decay begins at 7,200, this includes 200
constant-rate updates and 800 decay updates: 12.5\% of the complete training path.
The optimizer, data position, and depth RNG continue from the parent. In contrast,
the fixed-depth repair repeats only the final 10\% decay segment.

\begin{table}[htbp]
\centering
\caption{Primary shared-cache continuation losses for the variable-depth model, before
repair, with whole-run EMA-normalized consistency, and with late fixed-scale consistency
($\downarrow$). Numbers in parentheses are mean selected recurrent passes.}
\label{tab:app-adaptive}
\begin{tabular}{lrrr}
\toprule
Serving rule & Untreated & Whole-run consistency & Late consistency \\
\midrule
Fixed $R=3$ & 3.4187 (3.00) & 3.3899 (3.00) & 3.3862 (3.00) \\
Fixed $R=4$ & 3.4115 (4.00) & 3.3803 (4.00) & 3.3761 (4.00) \\
Fixed $R=6$ & 3.4149 (6.00) & 3.3832 (6.00) & 3.3779 (6.00) \\
Halting, cap 6 & 3.4151 (5.17) & 3.3826 (4.18) & 3.3816 (4.75) \\
\bottomrule
\end{tabular}
\end{table}

Halting compares the cosine of successive hidden states at each token, beginning after
the second pass, and uses threshold 0.98 with a cap of six. A stopped row keeps its
selected state and donor; batched computation continues until all rows stop or the cap
is reached. Mean selected depth is therefore not executed GPU work. Fixed $R=4$ has
lower loss and fewer selected passes than this halting rule in every model in
Table~\ref{tab:app-adaptive}. Halting-minus-$R4$ differences are
$0.0035\ (0.0003)$, $0.0023\ (0.0001)$, and $0.0055\ (0.0002)$, respectively.

Late consistency costs 0.0011--0.0017 in full-cache continuation loss and at most
0.0018 in chunk-standard loss. Its residual continuation penalty is $-0.0015$,
$0.0082$, and $0.0136$ at fixed depths 3, 4, and 6; under halting it is 0.0173.
For the halting comparison the full-cache reference uses fixed depth 6, so this
penalty includes both the cache policy and the stopping rule. The approximately
67\% reduction of that penalty should not be generalized from the fixed-depth
81--93\% range. The earlier global cosine rule also depends on evaluation mode:
in standard parallel evaluation it stops at four passes, whereas the shared pass
at threshold 0.98 and cap 6 has mean depth 4.86.

\section{Measured memory and decoding time}
\label{app:hardware}

Table~\ref{tab:app-hardware} measures an eager PyTorch decoder on one H100 after filling
its cache to 2,048 tokens by sequentially processing random token inputs from empty
caches. This fill differs from the ordinary parallel prompt prefill used for the
continuation-loss evaluation. For each configuration, the preceding decoder and cache are
released before resetting peak-memory statistics. Each timing is the median of three
trials of 256 continuation steps on an already filled cache; no subtraction of separate
prefill timings is used. Allocated-memory peaks include the cache fill and timed steps.
They exclude ordinary parallel-prefill workspace and are not end-to-end request
memory or latency measurements.
The last column subtracts the allocation measured immediately before constructing that
same batch-size configuration, not a base from a different batch size.

\begin{table}[htbp]
\centering
\caption{Isolated decoder measurements. Step times are milliseconds ($\downarrow$);
$B$ is batch size. Peak and peak-minus-base memory are MiB at $B=160$ ($\downarrow$).}
\label{tab:app-hardware}
\small
\setlength{\tabcolsep}{4pt}
\begin{tabular}{rlrrrrrr}
\toprule
Width & Layout & $B=1$ & $B=16$ & $B=64$ & $B=160$ & Peak & Peak$-$base \\
\midrule
768 & Cell, standard & 9.1 & 8.6 & 14.0 & 29.1 & 8259 & 7897 \\
768 & Cell, shared & 8.3 & 9.0 & 13.9 & 28.9 & 2980 & 2618 \\
768 & Untied & 7.8 & 7.7 & 13.5 & 28.7 & 8556 & 7996 \\
1024 & Cell, standard & 9.2 & 8.4 & 17.6 & 20.7 & 11136 & 10561 \\
1024 & Cell, shared & 8.3 & 8.4 & 17.3 & 37.2 & 4097 & 3522 \\
1024 & Untied & 8.0 & 8.2 & 17.3 & 20.6 & 11691 & 10740 \\
\bottomrule
\end{tabular}
\end{table}

At batch 160, full sharing reduces measured peak memory relative to the same cell's
standard layout by 2.77 times at width 768 and 2.72 times at width 1024. Relative to
the untied controls, the ratios are 2.87 and 2.85. These are total allocated peaks,
not just persistent KV storage. Small-batch step times range from approximately
7.7 to 9.2 ms. At width 1024 and batch 160, shared decoding is slower than standard
decoding (37.2 versus 20.7 ms). The present measurements do not isolate the cause
of this slowdown or predict an optimized serving engine's latency.

\begin{figure}[htbp]
\centering
\includegraphics[width=0.84\linewidth]{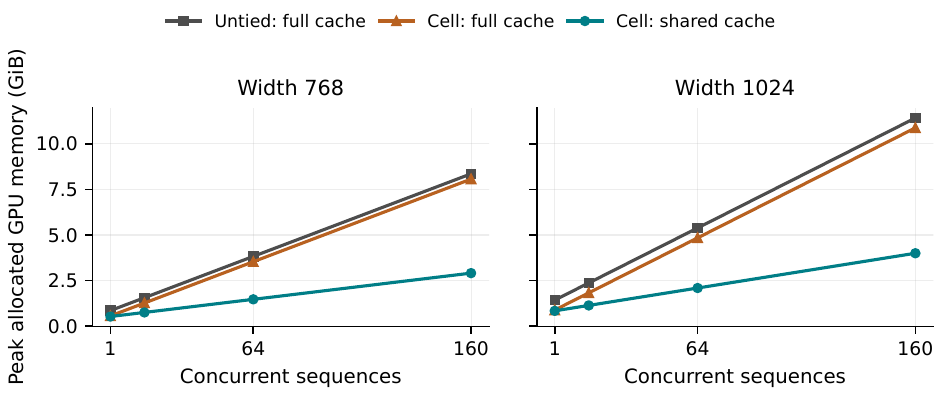}
\caption{Peak allocated memory by batch size for the isolated decoder measurements.
Persistent-KV allocation and total peaks are distinct quantities; model weights and
scratch computation contribute to the latter.}
\label{fig:app-memory}
\end{figure}

\FloatBarrier
\section{Prespecified comparisons and decision rules}
\label{app:prereg}

Table~\ref{tab:app-prespec} distinguishes comparisons specified before their results
from exploratory analyses. The consistency experiments followed discovery of the
shared-cache penalty, but their decision rules preceded the corresponding runs.
Coefficient selection, the cross-width parameter match and unreplicated ablations
remain exploratory; the held-out repair evaluation used a previously selected
coefficient. Passing a practical loss threshold does not imply zero degradation
or statistical equivalence.

\begin{table}[htbp]
\centering
\caption{Compact comparison and decision-rule record. Outcome statements apply only
to the tested scales and evaluation protocols.}
\label{tab:app-prespec}
\small
\begin{tabular}{p{0.22\linewidth}p{0.36\linewidth}p{0.32\linewidth}}
\toprule
Comparison & Rule or scope fixed beforehand & Observed outcome \\
\midrule
Untied control & Match reference forward FLOPs within 5\% before comparing losses.
& Fifteen blocks; residual forward mismatch 1.5--2.4\%. \\
Width/token ladder & Widths 384/512/768/1024, three annealed token endpoints, same data
and recipe. & Complete; recurrent gaps generally widen with tokens and shrink with width. \\
Gradient-path control & Cell/injection crossed with full/last-eight gradients;
width 512, uniform depth 4--32. & Large truncation penalty for the cell; small
full-gradient difference between updates. \\
Training-seed replication & Three paired seeds for the width-768 two-pass and untied
models, with fixed data order. & Shared-minus-untied gap positive in all nine endpoint/seed pairs. \\
Repair effect and quality & Shared continuation improvement at least 0.01 nats, with
paired uncertainty; chunk-standard degradation at most 0.005. & Selected late donor
repair meets both at tested widths and adaptive caps; whole-run adaptive treatment
exceeds the quality limit. \\
Branch control & Before comparing auxiliary terms, reproduce parent parallel losses
within 0.005 and continuation penalty within 0.01. & Control meets both tolerances. \\
Recipe selection and confirmation & Select among tested coefficients on original spans;
then evaluate the selected width-768 repair on fresh disjoint documents. & Both
training seeds retain the improvement, with small measured full-cache costs. \\
\bottomrule
\end{tabular}
\end{table}

\end{document}